%% file: main.tex
\documentclass[runningheads]{llncs}

\usepackage{eccv}

\usepackage{eccvabbrv}

\usepackage{graphicx}
\usepackage{booktabs}

\usepackage[accsupp]{axessibility}  %

\usepackage{hyperref}

\usepackage{orcidlink}

\usepackage{silence}
\usepackage[ruled]{algorithm2e}
\SetKwComment{Comment}{// }{}
\SetKwInput{KwInput}{\textbf{Input} }
\SetKwInOut{KwOutput}{\textbf{Output} }
\usepackage[nolist]{acronym}
\usepackage{csquotes}
\usepackage{bm}
\usepackage{booktabs}
\usepackage{multirow}
\usepackage{makecell}
\usepackage{seqsplit}
\usepackage{tikz}
\usetikzlibrary{3d}
\usetikzlibrary{pgfplots.groupplots}
\usepackage{pgfplots}
\pgfplotsset{compat=1.18}
\usepackage{shellesc} %
\usepackage{framed}

\def\twod{$2\textnormal{D}$\xspace}
\def\threed{$3\textnormal{D}$\xspace}
\def\wrt{w.r.t.\xspace}

\makeatletter
\renewcommand{\paragraph}{%
    \@startsection{paragraph}{4}%
    {\z@}{0.5\baselineskip}{-0.5em}%
    {\normalfont\normalsize\bfseries}%
}
\makeatother

\newcommand\inputpgf[2]{{
\let\pgfimageWithoutPath\pgfimage
\renewcommand{\pgfimage}[2][]{\pgfimageWithoutPath[##1]{#1/##2}}
\let\includegraphicsWithoutPath\includegraphics
\renewcommand{\includegraphics}[2][]{\includegraphicsWithoutPath[##1]{#1/##2}}
\input{#1/#2}
}}

\DeclareMathOperator{\atantwo}{atan2}

\graphicspath{{./img/}}

\begin{document}

\title{NIGHT - Non-Line-of-Sight Imaging from Indirect Time of Flight Data} 

\titlerunning{NIGHT - NLoS Imaging from iToF Data}

\author{Matteo Caligiuri\orcidlink{0009-0006-2928-1047} \and
Adriano Simonetto \and
Pietro Zanuttigh\orcidlink{0000-0002-9502-2389}}

\authorrunning{M.~Caligiuri et al.}

\institute{Università degli Studi di Padova - DEI, Via Gradenigo 6/b 35131, Padova, Italia \\
\email{\{matteo.caligiuri,zanuttigh\}@dei.unipd.it, sadriano95@gmail.com}
}

\maketitle

\begin{abstract}
  The acquisition of objects outside the Line-of-Sight of cameras is a very intriguing but also extremely challenging research topic. Recent works showed the feasibility of this idea exploiting transient imaging data produced by custom direct Time of Flight sensors. In this paper, for the first time, we tackle this problem using only data from an off-the-shelf indirect Time of Flight sensor without any further hardware requirement. We introduced a Deep Learning model able to re-frame the surfaces where light bounces happen as a virtual mirror. This modeling makes the task easier to handle and also facilitates the construction of annotated training data. From the obtained data it is possible to retrieve the depth information of the hidden scene. We also provide a first-in-its-kind synthetic dataset for the task and demonstrate the feasibility of the proposed idea over it.

\end{abstract}

\input{glossary}

\input{sec/01_intro}

\input{sec/02_related_work}

\input{sec/03_tof_intro}

\input{sec/04_proposed_method}

\input{sec/05_dataset}

\input{sec/06_experiments}

\input{sec/07_limitations}

\input{sec/08_conclusions}

\input{sec/09_acknowledgment}

\bibliographystyle{splncs04}
\bibliography{main}
\end{document}

%% file: glossary.tex
\begin{acronym}
    \acro{tof}[ToF]{Time of Flight}
    \acro{itof}[iToF]{in\-di\-rect Time of Flight}
    \acro{dtof}[dToF]{di\-rect Time of Flight}
    \acro{nlos}[NLoS]{Non-Line-of-Sight}
    \acro{cv}[CV]{Com\-put\-er Vi\-sion}
    \acro{brdf}[BRDF]{Bi\-di\-rec\-tio\-nal Re\-flec\-tan\-ce Dis\-tri\-bu\-tion Func\-tion}
    \acro{snr}[SNR]{Sig\-nal-to-Noise-Ra\-ti\-o}
    \acro{nir}[NIR]{Near-InfraRed}
    \acro{amcw}[AMCW]{Am\-pli\-tude Mo\-du\-la\-ted Con\-ti\-nu\-ous Wave}
    \acro{rtt}[RTT]{Round Trip Time}
    \acro{mpi}[MPI]{Mul\-ti\--\-Path In\-ter\-fe\-ren\-ce}
    \acro{nn}[NN]{Neu\-ral Net\-work}
    \acro{dl}[DL]{Deep Lear\-ning}
    \acro{ml}[ML]{Ma\-chine Learn\-ing}
    \acro{los}[LoS]{Line-of-Sight}
    \acro{spad}[SPAD]{Sin\-gle-pho\-ton avalanche diode}
    \acro{fov}[FoV]{Field of View}
    \acro{dlct}[D-LCT]{Di\-rec\-tion\-al Line-Cone trans\-for\-ma\-tion}
    \acro{nerf}[NeRF]{Neu\-ral Ra\-di\-ance Field}
    \acro{mlp}[MLP]{Mul\-ti-Lay\-er Per\-cep\-tron}
    \acro{netf}[NeTF]{Neural Transient Field}
    \acro{mc}[MC]{Monte Carlo}
    \acro{gt}[GT]{Ground Truth}
    \acro{dnn}[DNN]{Deep Neu\-ral Net\-work}
    \acro{mpi}[MPI]{Mul\-ti\--\-Path In\-ter\-fe\-ren\-ce}
    \acro{qvga}[QVGA]{Quar\-ter Vi\-de\-o Gra\-phic\-s Ar\-ray}
    \acro{cnn}[CNN]{Con\-vo\-lu\-tion\-al Neu\-ral Net\-work}
    \acro{cgi}[CGI]{Com\-pu\-ter-Ge\-ne\-ra\-ted Im\-a\-gery}
    \acro{fov}[FoV]{Field of View}
    \acro{mae}[MAE]{Mean Ab\-so\-lute Er\-ror}
    \acro{iou}[IoU]{In\-ter\-sec\-tion o\-ver Union}
    \acro{miou}[mIoU]{mean In\-ter\-sec\-tion o\-ver Union}
    \acro{ste}[STE]{Straight-Through Es\-ti\-ma\-tor}
    \acro{mse}[MSE]{Mean Square Er\-ror}
\end{acronym}

%% file: sec/01_intro.tex
\section{Introduction} \label{sec:intro}
    Many different strategies have been proposed for the acquisition of \threed data, but, regardless of the technology used, a key requirement to acquire \threed information from a real-world scene is to ensure that there is a direct path between the sensor and the acquired point. This setting is denoted as \acf{los} acquisition. Recent research showed that active devices could also be used for the extremely challenging task of \acf{nlos} imaging. This essentially concerns being able to capture a scene or an object located outside the direct \acf{fov} of the sensor. This category of imaging goes one step further \wrt its \ac{los} counterpart, analyzing the light scattered from multiple surfaces along indirect paths to reveal the \threed shape of objects located outside the \ac{los} \cite{faccio2020non}. \ac{nlos} imaging using a \threed sensor is currently feasible only in very few scenarios where there is a simple way for the emitted light to reach the hidden scene. In particular, the \textit{looking behind a corner} setting depicted in \cref{subfig:mirror-trick:base-scene} can be considered the baseline for this task. 
    The goal of our work is thus to use an \acf{itof} sensor to perform \ac{nlos} \threed imaging in this setup. More precisely, the objective is to build a learned approach that, given the raw \ac{itof} measurements as input, can retrieve the depth map of an object located behind a corner, outside of the direct \ac{fov} of the camera. We have chosen to use an \ac{itof} instead of \acf{dtof} cameras, used in other works tackling this setting, because it guarantees a greater lateral resolution \cite{itof2dtof}. Furthermore, this kind of device is able to retrieve a more refined depth \wrt the one measured by a \ac{dtof}. 
    Notice also that most current \ac{nlos} works require custom cameras with impressive specifications while we target the usage of off-the-shelf consumer camera. This makes it more suitable for integration into other devices such as cars, robots and helmets. 
    As of the time of writing, there is a limited number of approaches tackling \ac{nlos} imaging with \ac{itof} data and, consequently, a dedicated dataset for this specific application does not currently exist. For this reason, we have decided to build it from scratch. 
    
    The proposed approach exploits the \textit{Mirror Trick}, an innovative idea that we have introduced to reframe the problem as a \ac{los} one by interpreting the front wall as a virtual mirror, thus simplifying the definition of the ground truth in the dataset and the task itself. Note that the reflecting wall in the dataset could be either diffuse or glossy and the \textit{mirror trick} is used only for re-framing the task. The data are then used to train a deep learning model that predicts sensor data in the \textit{mirrored} setting and thus extracts depth information for the original task from the \ac{itof} data. The main contributions of this work are:
    \begin{itemize}
        \item \textbf{mirror trick}: A new approach to reframe a \acs{nlos} setting as a \acs{los} one;
        \item \textbf{deep learning model}: A \ac{dnn} able to estimate the depth in this setting using  combined depth and shape losses;
        \item \textbf{\acs{itof}  dataset}: The first synthetic dataset with \acs{itof} data for \acs{nlos} imaging.
    \end{itemize}

%% file: sec/02_related_work.tex
\section{Related work} \label{sec:related}
    \ac{nlos} perception using depth sensors is a new field, but there are already some works that pursue this goal \cite{2019chenSteadyStateNonLineOfSightImaging, 2019lindellAcousticNonLineOfSightImaging, 2019maedaThermalNonLineofSightImaging, 2019saundersComputationalPeriscopyOrdinary, 2020scheinerSeeingStreetCorners, 2020tanakaPolarizedNonLineofSightImaging}. In this section, we focus only on the solutions based on the use of \ac{tof} sensors since these are closer to the proposed one. In particular, if we consider \ac{itof} sensors, we realize that only a few examples exist \cite{kadambiOccludedImagingTimeofFlight2016}. Furthermore, all of them assume to know some a priori information on the scene or to be in a simplified scenario. To our knowledge, we are the first to perform \ac{nlos} imaging using an off-the-shelves \acl{itof} sensor that illuminates in a single shoot (\textit{full field}) a scene on which we know anything else but the fact that it satisfies the \textit{behind a corner} setup. Indeed, the vast majority of existing works exploit data from \ac{spad} sensors embedded into \acl{dtof} cameras \cite{faccio2020non, xin2019theory, heide2014diffuse, young2020non, kirmani2009looking, velten2012recovering, buttafava2015non, shen2021non, grau2020deep, wang2023non, zhu2023compressive, heideDiffuseMirrors3D2014, lindellWavebasedNonlineofsightImaging2019, liuNonlineofsightImagingUsing2019, otooleConfocalNonlineofsightImaging2018}. This makes the task easier \wrt our setting since the direct and indirect light components are much simpler to separate. Another common trend in this field is to illuminate the scene not in a single shoot with a \textit{full filed illumination} but, rather illuminate a grid of points on the front wall one by one. This makes the reconstruction of the hidden scene much simpler since each single illumination step will not interfere with the others. On the other hand, the acquisition procedure becomes very long.

    \paragraph{Analytical methods}
    To the best of our knowledge, the only work exploiting \ac{itof} data is  \cite{kadambiOccludedImagingTimeofFlight2016}. This paper considers a huge simplification in the \textit{look behind a corner} setup by spatially splitting the \ac{tof} device into two separate entities: the light emitter, located in the \ac{nlos} and the actual sensor, located in the \ac{los}.
    A key work in the field is \cite{xin2019theory}, which introduces a novel approach to transient-based \ac{nlos} imaging. The authors present the concept of the \textit{Fermat path}, the specific photon path between the \ac{los} and \ac{nlos} scene, and develop the \textit{Fermat flow} algorithm for shape recovery.
    Heide \etal \cite{heide2014diffuse} exploits the fact that the temporal structure of the light, between a diffuse wall and a mirror, is left intact to recover objects outside the \ac{los}. They formulated the reconstruction task as a linear inverse problem.
    A joint albedo-normal approach to \ac{nlos} surface reconstruction using the \ac{dlct} is proposed in \cite{young2020non}.
    Kirmani \etal \cite{kirmani2009looking} proposed a new framework that uses transient data analysis to extract the properties of the \ac{nlos} scene.
    A combination of \ac{tof} data and computational reconstruction algorithms is used in \cite{velten2012recovering} to untangle the image information mixed by diffuse reflections.
    A \ac{nlos} imaging system that uses a single \ac{spad} pixel to collect \ac{tof} information is proposed in \cite{buttafava2015non}. This work focuses attention on practical aspects like power requirements, form factor, cost, and reconstruction time.
          
    \paragraph{Deep Learning based methods}
    Shen \etal \cite{shen2021non} fuse \ac{nlos} reconstruction with \ac{nerf} by developing a \ac{mlp} to represent a spherical volume \ac{netf}.
    A tailored autoencoder architecture, trained end-to-end, that maps \ac{nlos} transient images directly to a depth map is proposed in \cite{grau2020deep}. 
    The results are computed over a synthetic simulation of \ac{spad} measurements.
    Wang \etal \cite{wang2023non} proposed a general learning-based pipeline for \ac{nlos} imaging using only a few acquisition points. A \ac{nn} is tailored to learn the operator that recovers the high spatial resolution signal.
    A synthetic photon \ac{tof} histogram with picosecond temporal resolution is fed to a compressive imaging algorithm that exploits \ac{dl} in \cite{zhu2023compressive}.

%% file: sec/03_tof_intro.tex
\section{Basic principles on Time of Flight imaging} \label{sec:tof-intro}
    Before presenting ours method, we review the basic principles of \ac{tof} sensors, which fall into two main categories: \textit{\acf{itof}} and \textit{\acf{dtof}}. Regardless of the \ac{tof} technology, the measured data can be represented in various formats; we have chosen the \textit{phasorial} format.

    \subsection{Indirect Time of Flight} \label{sub:tof-intro:itof}
        Indirect Time of Flight sensors do not measure the distance directly from the time of flight along the light ray but instead, calculate it indirectly from the phase shift $\phi$ between the emitted modulated signal and the received one \cite{zanuttigh2016time}. Ideally, the camera is assumed to be composed of two co-located elements: the emitter (light source) and the sensor (detector). Notice that in a real device, they cannot be exactly in the same location and it is necessary to perform some additional compensation steps. The source emits an amplitude-modulated light signal ($o$) that travels toward the scene and is reflected back to the detector. The detector convolves the received signal ($r$) with a square signal characterized by the same frequency ($f_\textnormal{m}$) of the emitted one. By sampling this convolution on $4$ points, it is possible to extract the phase shift ($\phi$) between the two signals and use it to compute the distance. \ac{tof} devices based on this principle are also known as \ac{amcw} \ac{tof}. From the convolution samples, it is then possible to compute the amplitude ($A$) of the reflected signal and the phase shift ($\phi$) \cite{zanuttigh2016time}. In particular, using the phasor notation we get:
        \begin{equation} \label{eq:itof-phasor}
            c = Ae^{j\phi} = A \cdot (\cos(\phi) + j \sin(\phi)) = \Re + j \Im,
        \end{equation}
        the amplitude and phase correspond to:
        \begin{equation} \label{eq:amplitude-phi-phasor}
            A = \sqrt{\Im(c)^2 + \Re(c)^2},
            \qquad \;
            \phi = \atantwo \left( \Im(c),\ \Re(c) \right).
        \end{equation}
        The $\atantwo (\cdot)$ function is used to capture the full $2\pi$ range of the phase.
        At this point, using the phase ($\phi$), the modulation frequency ($f_\textnormal{m}$) and the speed of light ($c$), we can simply compute the target point distance ($d$) as:
        \begin{equation} \label{eq:distance}
            d = \frac{c}{4 \pi f_\textnormal{m}} \phi.
        \end{equation}

        For a better comprehension of the data generated by an \ac{itof}, it is essential to remember that the amplitude ($A$) directly correlates with the intensity of the incident light beam striking an object. The phase shift ($\phi$), conversely, is proportional to the depth of the object that reflected such a light ray.

    \subsection{Direct Time of Flight Cameras} \label{sub:tof-intro:dtof}
\begin{figure}[ht]
            \centering
            \resizebox{.8\textwidth}{!}{%
            \input{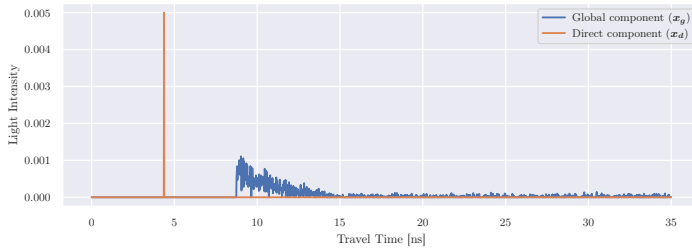}}
            \caption{Transient vector representation}
            \label{fig:backscattering}
        \end{figure}

        Direct Time of Flight cameras work by generating a \ac{nir} light pulse of known duration and then discretizing the front of the reflected light. The discretization is performed before the return of the whole light pulse using a fast camera shutter. The resulting portion of the reflected light signal is the element that describes the observed object. Using this approach, the depth is directly linked to the time delay of the received signal \cite{lefloch2013technical}. Indeed, knowing the light pulse duration ($t_\textnormal{pulse}$) and the shutter time ($\delta_s$), it can be computed as:
        \begin{equation} \label{eq:dtof-depth}
            d = \frac{c}{2  (t_\textnormal{pulse} + \delta_s)}.
        \end{equation}
        The output produced by this kind of sensor is sensibly different from the one of an \ac{itof}. A \ac{dtof} can return for each sensor pixel a \textit{transient vector} ($\bm{x}$) that describes the light intensity received at each time instant. This, as shown in \cref{fig:backscattering}, can be easily separated into its \textit{direct} ($\bm{x_d}$) and \textit{global} ($\bm{x_g}$) components, a challenging task for \ac{itof} data. This difference with its indirect counterpart is the key aspect that makes performing \ac{nlos} imaging easier using a \ac{dtof} sensor.
        Finally, note that converting a \ac{dtof} output to an \ac{itof} one is a straightforward analytical approach (see \cite{burattoDeepLearningTransient2021a} for the mathematical details) while the reverse is an ill-posed problem. Starting from the \ac{dtof} transient vector ($\bm{x}$) and the \ac{itof} measurement's model ($\bm{\Phi}$), the \ac{itof} data can be obtained as $\bm{c_\textnormal{\textbf{mf}}} = \bm{\Phi} \bm{x}$.
        
    \subsection{NLoS perception using ToF cameras} \label{sub:tof-intro:nlos}
        Regarding the task of \acl{nlos} imaging using a \ac{tof} sensor, the main idea is to analyze the output of a \ac{dtof} and from that extract the \textit{global component} ($\bm{x_g}$) excluding the direct one, since an object located in \ac{nlos}, for sure, will not be the closest object to the sensor. It is important to note that even if we can precisely extract the global component, discerning which element corresponds to noise, or to other reflections in the scene, and which to the hidden object is an extremely difficult and ill-posed task. Currently, most of the state-of-the-art approaches perform the extraction of useful data from ($\bm{x_g}$) using some kind of \ac{dl} model. Furthermore as stated in \cite{kadambiOccludedImagingTimeofFlight2016} one of the biggest limitations that arise from the use of off-the-shelves \ac{tof} device for \ac{nlos} imaging is the fact that the information coming from the hidden scene is carried by light rays that have performed at least three reflections. Each of them starts from the emitter, bounces on the front wall, then on the hidden object, then again on the front wall and finally, it reaches the sensor.
        Note that this represents the most ideal path since, in most cases, there will be more reflections before the light rays reach the sensor. This means that, even in the best scenario, such information necessarily travels a long path and so, the measured intensity greatly reduces, making the process of discerning between noise and useful information even harder. This problem is even more severe if we consider that the reflections coming directly from the front wall are extremely more intense and could saturate the sensor.
        As it is clear from what was said above, the standard approach is to extract meaningful information about the \ac{nlos} scene from the \textit{global component}. Using an \ac{itof} instead of a \ac{dtof} makes the task more complex. Indeed, as stated in \cref{sub:tof-intro:dtof}, the \ac{itof} output doesn't explicitly distinguish between direct and global components, making it harder to extract information from the hidden scene.
        It can be inferred from this general introduction that \ac{nlos} applications using \ac{tof} technology are restricted to functioning in just a few specific scenarios, such as \textit{look behind a corner}, due to the necessity that the light reflected from the hidden objects reach the sensor with a relatively simple path.
        
        \begin{figure}[ht]
            \centering
            \subfloat[Basic scene (standard)]{
                \includegraphics[width=.2\linewidth]{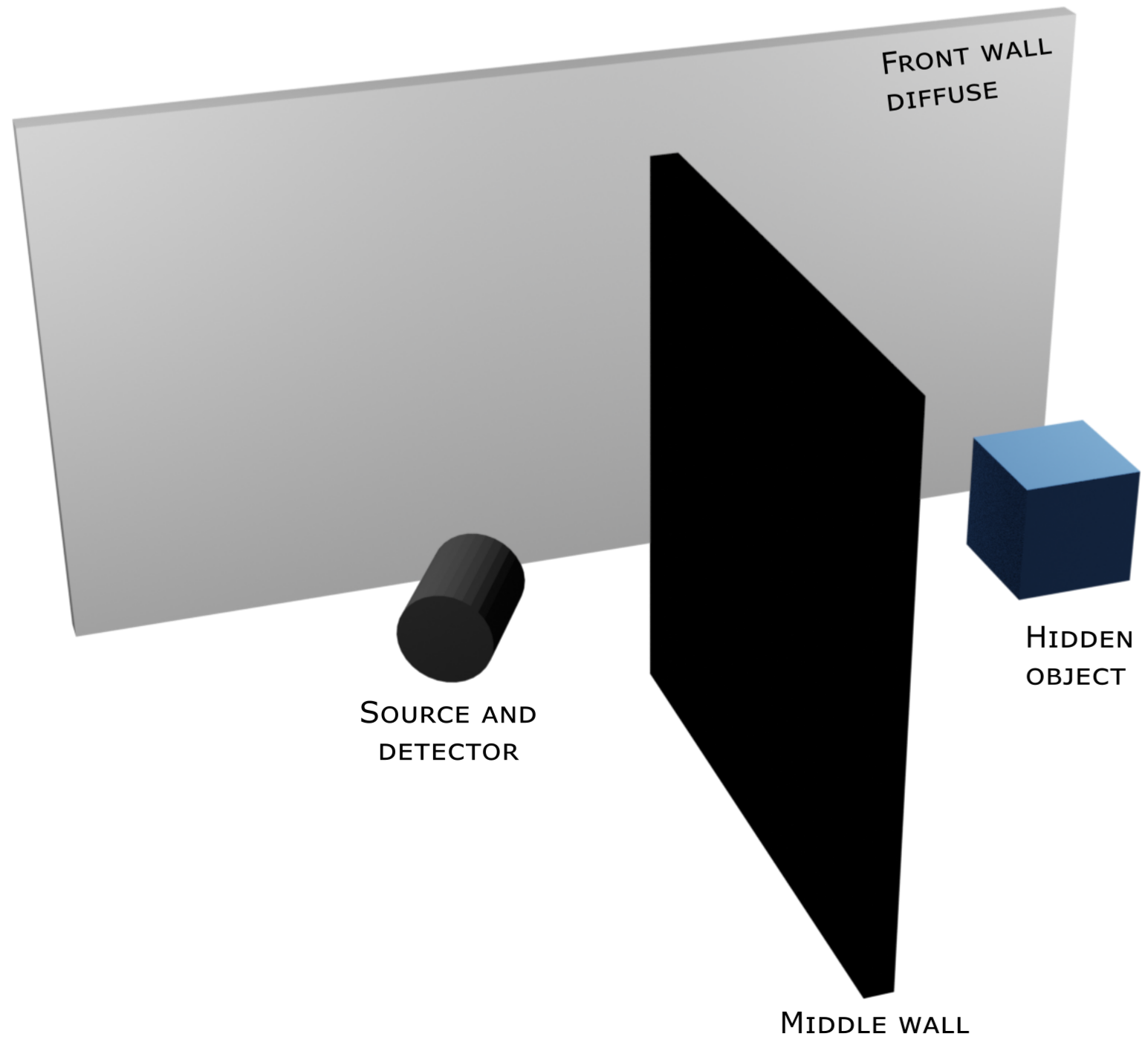}
                \label{subfig:mirror-trick:base-scene}
                }
            \hfil
            \subfloat[Basic scene (perfect mirror)]{
                \includegraphics[width=.2\linewidth]{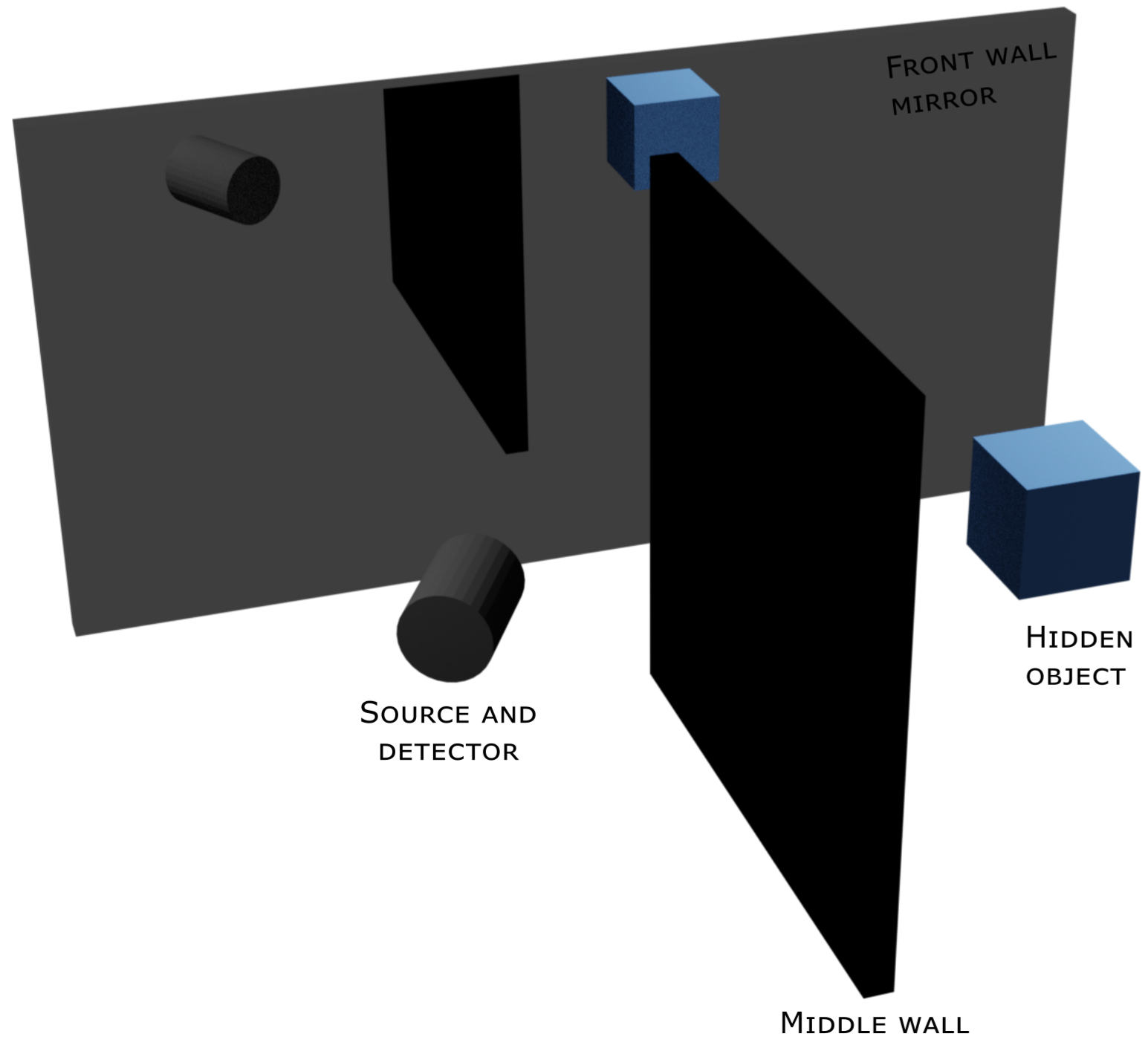}
                \label{subfig:mirror-trick:mirror-scene}
                }
            \hfil
            \subfloat[Basic scene (after \textit{Mirror Trick})]{
                \includegraphics[width=.2\linewidth]{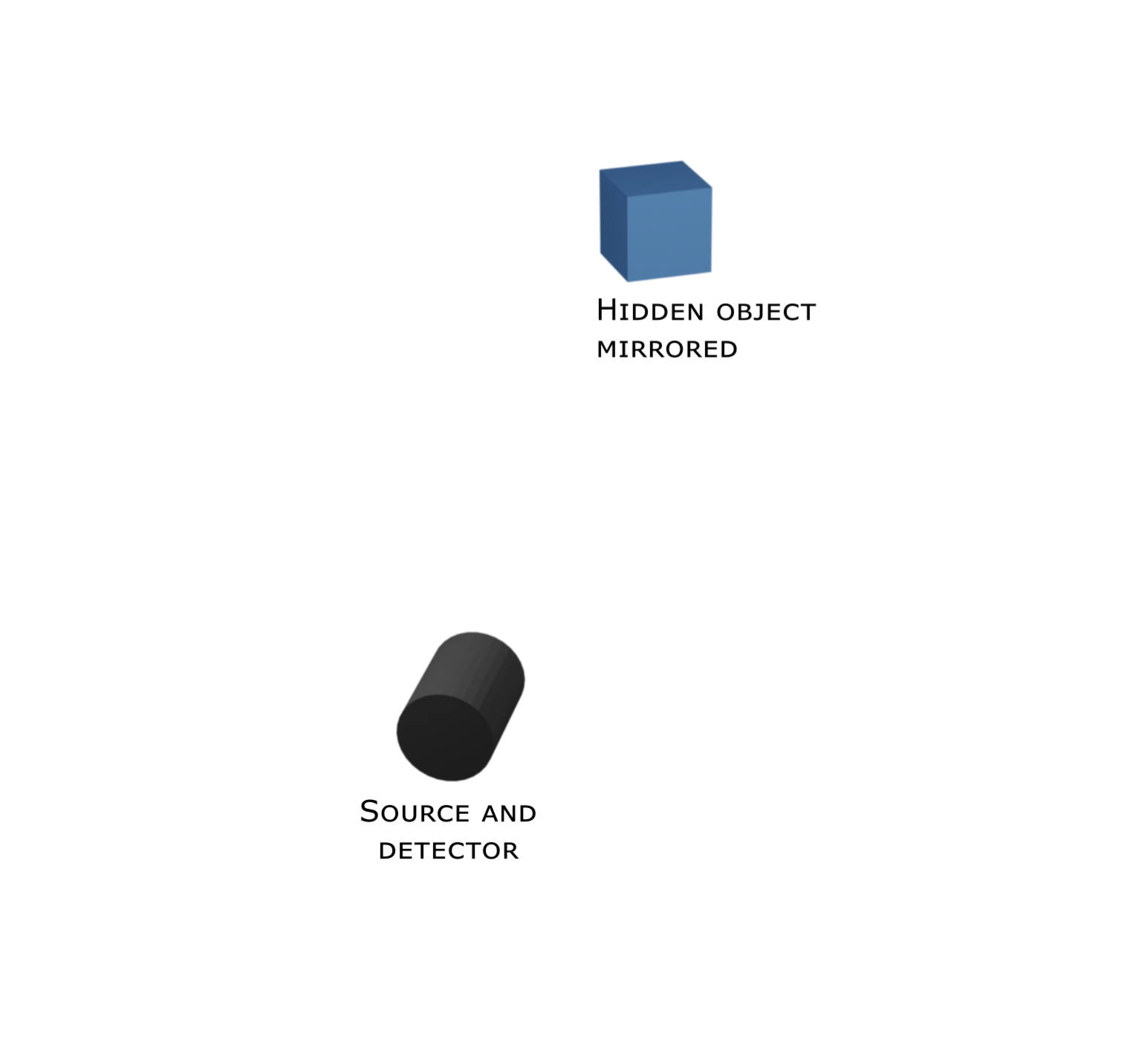}
                \label{subfig:mirror-trick:obj-scene}
                }
            \hfil
            \subfloat[Axis orientation]{
                \includegraphics[width=.2\linewidth]{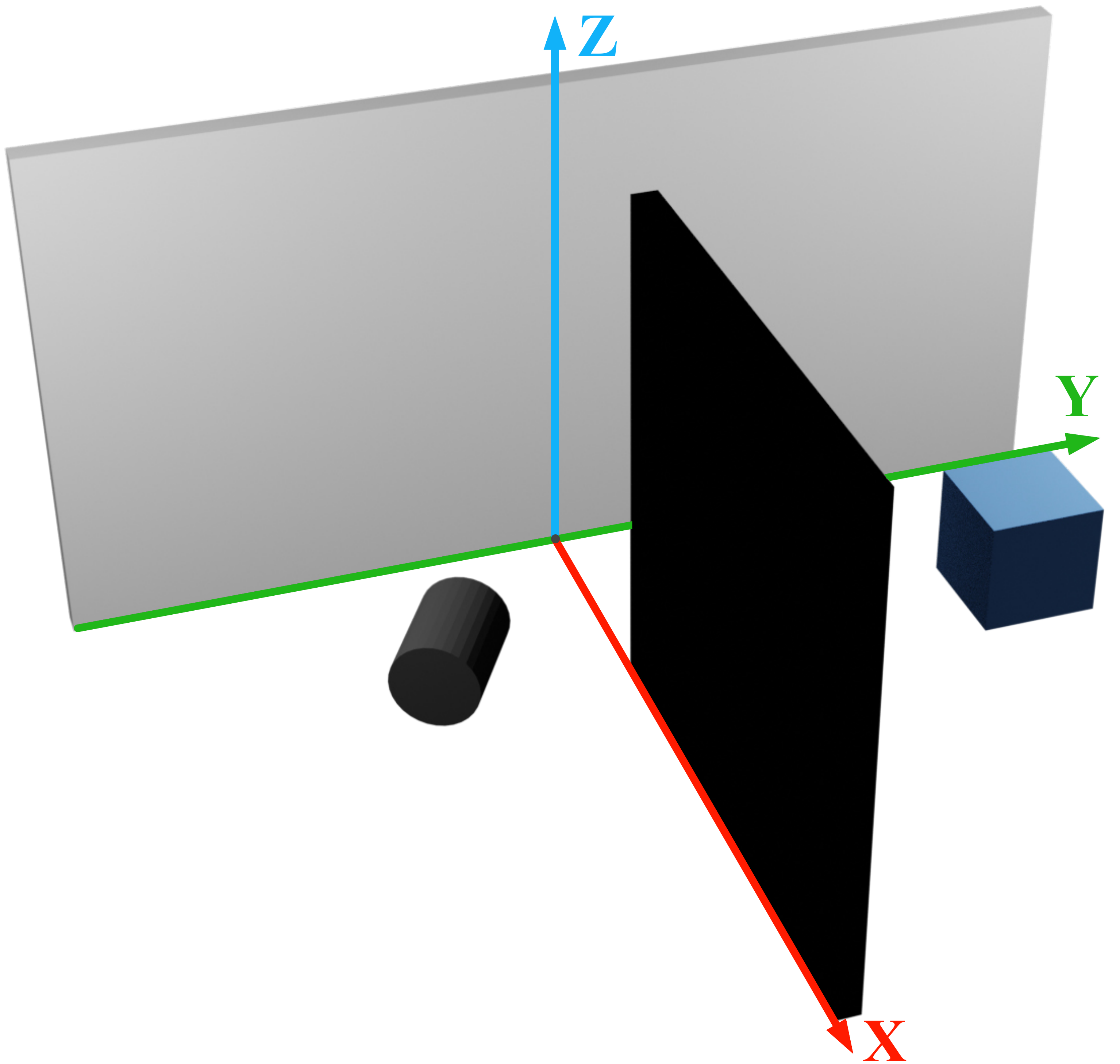}
                \label{subfig:mirror-trick:axis}
                }
            \caption{Representation of the \textit{Mirror trick}.}
            \label{fig:mirror-trick}
        \end{figure}

%% file: sec/04_proposed_method.tex
\section{Proposed method} \label{sec:method}
    As discussed in \cref{sec:intro}, the proposed \acl{nlos} imaging approach utilizes a \acl{dl} model based on \ac{itof} data. A straightforward method involves a \textit{supervised learning} model that takes \ac{itof} data as input to predict the hidden object's geometry. The network would directly output the depth map of the \ac{nlos} scene, as if viewed from the front wall. However, modeling the complex input-output relationship this way often leads to poor results. An alternative is to train the network to estimate \ac{itof} data for a \ac{los} scene, simplifying the task by avoiding direct \ac{tof}-to-depth conversion. Despite this, accurately simulating \ac{itof} data for an unknown scene remains challenging. Therefore, we propose a novel reformulation of the problem to make it more tractable for the learning model and simplify ground truth generation for training.
    
    \subsection{The mirror trick} \label{sub:method:mirror-trick}
        As previously introduced, we aim at simplifying the task that the network model has to perform, reducing as much as possible the implicit transformations. To this aim, we introduce the \textit{Mirror trick}. If we consider the toy scenario in which the front wall is a mirror, the \ac{nlos} settings do not hold anymore and the retrieving of the hidden object becomes straightforward. Exploiting some fundamental principles of optics, we can effortlessly explain this behavior. When the front wall functions as an ideal mirror, the reflections of incoming rays are perfectly specular, leading to a substantial simplification in the geometric description of the scene. With these specular reflections, we can now treat the trajectory traversed by the ray emitted from the sensor and its subsequent reflection off the front wall as a solitary ray with a length equivalent to the sum of the individual ray lengths between the sensor and the front wall and between the front wall and the object. A similar analysis applies to the returning path (from the object to the front wall and then to the sensor). The described setup is depicted in \cref{subfig:mirror-trick:mirror-scene}. Here, it is possible to see that the \ac{tof} sensor could directly see the object in the \ac{nlos} area as if it is in \ac{los}. Furthermore, if the mirror used is ideal, given how a \ac{tof} camera works, the depth map produced by the sensor in a setup like the one of \cref{subfig:mirror-trick:mirror-scene}, is the same as that produced from the setup of \cref{subfig:mirror-trick:obj-scene}. Essentially, if the front wall is an ideal mirror it is possible to assume that the \ac{nlos} object appears to the sensor as if it is flipped around the front wall plane. The \textit{mirror trick} also allows our method to be completely agnostic to the material of which the front wall is composed. Indeed, if we found a way to teach a \ac{dl} model how to convert a surface of unknown material to something as close as possible to an ideal mirror, we will be able to retrieve information about the \ac{nlos} scene with a small effort. To summarize, the \textit{mirror trick} is a clever approach to handle \textit{look behind a corner} setup without any knowledge about the front wall material.
        In particular, a learning model exploiting it should be able to take in input \ac{itof} measurements of a scene like the one of \cref{subfig:mirror-trick:base-scene} and transform them in such a way that the obtained data still represents \ac{itof} measurements data but acquired in a setup like the one depicted by \cref{subfig:mirror-trick:mirror-scene}.
        Once we have reached this goal, the obtained data is the same as the one coming from a scene like the one in \cref{subfig:mirror-trick:obj-scene} thus extracting the \ac{nlos} scene becomes a straightforward operation. We want to stress that the application of the \textit{mirror trick} does not require the front wall to be a mirror, but is used to convert a general \textit{look behind a corner} scene, with no further assumption on the used materials, to one that uses an ideal mirror as a reflective wall. Another benefit of using the \textit{mirror trick} is that it simplifies the \ac{gt} data generation process. This represents a key step of our work, since generating good quality data, that can be reliably used as \acl{gt}, is extremely challenging in this field. The idea behind the ground truth data generation is very similar to the one presented in the paragraph above: it can be directly generated by assuming that the front wall is an ideal mirror. 
        Under this assumption to generate each \ac{gt} sample, it is sufficient to take each scene in the standard setting (\cref{subfig:mirror-trick:base-scene}), remove the front wall, and flip the hidden object across the wall plane (\cref{subfig:mirror-trick:obj-scene}).
        After applying the \textit{mirror trick}, we simulate the \ac{tof} data for the obtained scene by feeding it into the \ac{tof} simulator (\texttt{Mitsuba 2 Renderer} \cite{NimierDavidVicini2019Mitsuba2}), as explained in \cref{sec:dataset}.

    \subsection{Deep learning framework} \label{sub:method:network}
        \paragraph{Input data}
        The input is defined as raw \ac{itof} measurements of a \textit{look behind a corner} scene like the one in \cref{subfig:mirror-trick:base-scene}. We decided to represent the data in phasor format so that each measurement has a real ($\Re(c)$) and an imaginary ($\Im(c)$) component. In order to provide the network with enough data, and following the same rationale presented in \cite{2019agrestiUnsupervisedDomainAdaptation,itof2dtof} we have decided to perform each measurement at three different acquisition frequencies: $20\textnormal{MHz}$, $50\textnormal{MHz}$ and $60\textnormal{MHz}$. Doing that, as stated also in \cite{burattoDeepLearningTransient2021a} it is possible to get different interfering characteristics between the visible wall and the hidden object. To improve stability to such large input values, a normalization step is also required: we divided each component of the \ac{itof} measurement by the corresponding amplitude at $20\textnormal{MHz}$, calculated using \cref{eq:amplitude-phi-phasor}. By doing that we ensure that the value range of the input data is $[0,\ 1]$. A side effect of this operation is the removal of the information about the amplitude at $20\textnormal{MHz}$, which is useful for extracting information about the noise level of the measurements. To overcome this issue, we have decided to add this information (divided by a constant to bring it also in the $[0,\ 1]$ range) as an additional channel to the normalized \ac{itof} data. After this small pre-processing, we feed the network with a matrix of dimension: $b_s \times 7 \times w \times h$. Where $b_s$ is the batch size, while $w$ and $h$ are the image dimensions. The second dimension contains the $20\textnormal{MHz}$ amplitude followed by the three normalized real components and the three normalized imaginary components (one for each frequency).

        \paragraph{Data augmentation}
        In order to maximize the amount of information that the \ac{dnn} can gather from the input data, we performed an augmentation step of the training data. More precisely we applied a random rotation in $[\pm 180]^\circ$, a random translation in $[\pm 0.2 \cdot (w, h)]$px, flipped samples vertically and horizontally and added Gaussian noise with mean $\mu = 0.0$ and standard deviation $\sigma = 1.0$.

        \paragraph{Output data}
        As for the output, we train the network to predict the \ac{itof} data, only at $20\textnormal{MHz}$, corresponding to the scene obtained by applying the \textit{mirror trick} to the input (\cref{subfig:mirror-trick:obj-scene}). 
        This means that the network produces an output \ac{itof} measurements of the input scene where the front wall is transformed into an ideal mirror. In this way, we reduce to the minimum the complexity of the task the network has to perform internally, and at the same time, we obtain an output from which it is straightforward to compute the depth map of the hidden object. Indeed, from the network output, it is enough to use \cref{eq:amplitude-phi-phasor} together with \cref{eq:distance} to obtain the desired depth map. For coherence with the input \ac{itof} samples, the output ones are expressed in phasor form ($\Re(c)$ and $\Im(c)$).

        \begin{figure}[ht]
            \centering
	        \includegraphics[width=1\linewidth]{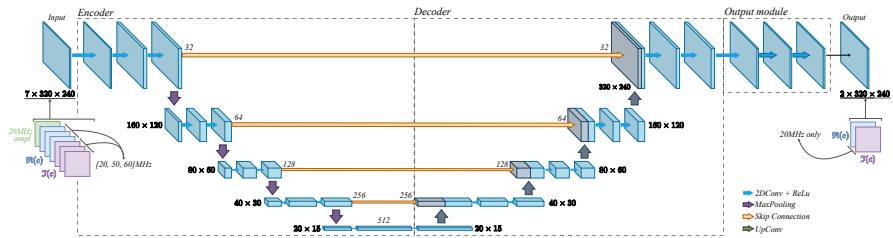}
	        \caption{Schematized representation of the proposed \ac{nn} model}
            \label{fig:net-model}
	    \end{figure}

        \paragraph{\acs{nn} architecture}
        We work with input data that can be seen as a multi-channel image (each layer of the input data is a \twod matrix where each pixel contains an intensity value) from which we want to retrieve another image-like representation, similarly to a denoising or image translation problem. For this reason, the use of a \twod convolutional model is a fairly straightforward choice. In particular, as it is possible to see from \cref{fig:net-model}, we used an Encoder-Decoder architecture similar to a \textit{U-Net} \cite{ronneberger2015u}. This model has been widely used for segmentation and denoising tasks and it has good convergence properties even with a limited amount of data. This is a key requirement in our case since there is no existing dataset and we could not build a very large one. Finally, we concatenated to the output of the Encoder-Decoder section a set of standard \twod convolutional layers (with ReLU activation function) to fine-tune the prediction and reduce the number of output channels to $2$, \ie, the two components of the output \ac{itof} data.

        \paragraph{Loss function}
        The loss function of the proposed model aims at jointly optimizing the shape of the hidden object and the depth value of its points. To this end, the loss function is designed as a combination of two terms:
        \begin{equation} \label{eq:lossv2}
            L_{\textnormal{MAE}} = w_l L_{\textnormal{MAE}}^{obj}\!+\! L_{\textnormal{MAE}}^{bgr},\quad
            L_{\textnormal{IoU}} = 1 \!- \!IoU(s_p, s_t);\quad 
            L = \frac{L_{\textnormal{MAE}} \!+\! L_{\textnormal{IoU}}}{2}.
        \end{equation}
        The first term, $L_{\textnormal{MAE}}$, is a modified version of the standard \ac{mae} loss function on the depth values. Since in our setting, there are many more pixels corresponding to the background than corresponding to hidden objects, we balanced the contribution of the object and background in the \ac{mae} loss in order to give a similar relevance to both regions. We separately calculated the \ac{mae} loss on the object region $L_{\textnormal{MAE}}^{obj}$ and the same loss on the background $L_{\textnormal{MAE}}^{bgr}$ and then computed the loss as a weighted sum of the two. 
        In particular, the weighting parameter is computed as $w_l = \alpha \cdot \frac{|B|}{|O|}$ where $\alpha$ is empirically set to $1/7$ while $|B|$ and $|O|$ are, respectively, the number of pixels in the background and object regions. The second term $L_{\textnormal{IoU}}$, is computed using the \ac{iou} metric across the predicted and ground truth segmentation maps $s_p$ and $s_t$, to help the \ac{nn} model distinguish between the object and background regions (foreground/background separation) and ultimately helps the algorithm to better reconstruct the precise shape of the hidden object. The computation of the segmentation maps is done by thresholding the depth values, a non-differentiable operation. This represents a challenge for the training procedure since after that, it is no longer possible to backpropagate the gradient. To solve this issue, we implement this thresholding function together with a \ac{ste} \cite{yin2019understanding} in order to perform the non-differentiable operation only during the forward step, while in the backward step, we perform a simple linear, and so differentiable operation. Before implementing the \ac{iou} together with the \ac{ste}, we also tested the Lováz-Softmax loss \cite{berman2018lovasz} as an alternative method for direct optimization of the \ac{iou}, but this loss led to lower performances.
        
        \paragraph{Training parameters}
        The training procedure was carried out for $695$ epochs on the $2833$ training samples obtained from the proposed dataset after applying data augmentation. It took $\approx 4$h and $40$min using the Adam optimizer with a learning rate of $lr = 1 \cdot 10^{-4}$. The training was set to run for $1500$ epochs, but we applied an early stopping criterion, i.e., the learning procedure was automatically stopped after $150$ consecutive epochs with no validation error improvement. The model was trained on a \textit{NVIDIA RTX 2080Ti} with $11$GB of VRAM.

%% file: sec/05_dataset.tex
\section{Proposed dataset} \label{sec:dataset}
    As noted in \cref{sec:intro,sec:related}, using \ac{itof} sensors for \ac{nlos} imaging is relatively unexplored. Therefore, we created a synthetic dataset from scratch to train \ac{nn}s due to the complexity of real-world acquisition. This synthetic dataset allows precise control over acquisition parameters. Although real-world experiments are crucial, demonstrating feasibility with synthetic data is a reasonable first step. To transition to real data, a strategy would be to pre-train on the synthetic dataset and fine-tune with real data. Thus, our dataset provides a solid foundation for future work. A typical dataset scene is shown in \Cref{subfig:mirror-trick:base-scene}, offering a simple yet useful starting point.

    \subsection{Rendering pipeline} \label{sub:dataset:rendering-pipeline}
        To generate the dataset, we automated the creation of each \threed scene and used a rendering engine to simulate observations by a \acl{tof} sensor. We built an automatic pipeline in \texttt{Blender 2.83} \cite{blender} to generate scene variations starting from a set of primitives. These scenes were converted to the required format using the \texttt{Mitsuba Blender Add-on} \cite{mitsuba2-blender}. We employed the \texttt{Mitsuba2 renderer} \cite{NimierDavidVicini2019Mitsuba2}, an open-source, highly realistic renderer, specifically using \texttt{Mitsuba2-transient-nlos} \cite{royo2022non} to render the scene as a \textit{transient vector}. This renderer's \textit{ray tracing} technology accurately simulates the \ac{brdf} of all surfaces, ensuring realistic scene rendering. Finally, we converted the data from \ac{dtof} to \ac{itof}.

    \subsection{Structure of the dataset} \label{sub:dataset:structure}
        We generated a dataset composed of $1344$ scenes: each scene is acquired using $10^6$ rendering samples, $2000$ temporal bins of size $0.01$s (that is equivalent to $0.01$m), a sensor horizontal \ac{fov} of $60^{\circ}$ and a resolution of $320 \times 240$ pixels (\acs{qvga}). Another key aspect is the fact that the illumination is \textit{full-field}, i.e., the whole scene is acquired and illuminated in a single step. This means that if we capture the same scene in the real-world the acquisition process requires a single shoot (allowing real-time). This last characteristic highly differentiates our model from those based on \ac{spad} sensors, since they are limited to a sparse acquisition performed one dot at a time (much longer acquisition time). 
        An essential aspect of a synthetic dataset is that it has to guarantee a high enough level of variability. This is fundamental since the data contained in the dataset must contain enough information to allow a \ac{nn} model to learn the desired task. In order to ensure the aforementioned variability, we decided to consider different locations and orientations for the hidden objects and also to allow the front wall to assume different roughness values. We consider a set of primitive geometrical shapes for objects, \ie, the ones presented in \cref{fig:hidden-objects} with their locations, orientation and dimensions. Possible positions and rotations are randomly sampled. In addition to those basic shapes, $\frac{1}{7}$ of the scenes contains, as a hidden object, a random combination of two primitives sampled randomly. The distance between the two walls is fixed at $70$cm. Finally, the roughness value of the front wall can take values in $\alpha \in \lbrack 0.3,\ 1 \rbrack$ with a step $0.05$. This corresponds to a variation in the glossiness of the wall, which implies that our dataset does not contain only diffuse surfaces. Code and dataset are available \href{https://github.com/LTTM/NIGHT-nlos-imaging-from-itof-data}{here}\footnote{\href{https://github.com/LTTM/NIGHT-nlos-imaging-from-itof-data}{https://github.com/LTTM/NIGHT-nlos-imaging-from-itof-data}}.
        
        \begin{figure}[t]
            \centering
            \begin{minipage}{.42\textwidth}
                \centering
                \begin{framed}
                    \input{img/section_05/objects}
                \end{framed}
            \end{minipage}
            \quad
            \begin{minipage}{.31\textwidth}
                \centering
                \resizebox{1\textwidth}{!}{%
                \input{tab/dataset_tab}}
            \end{minipage}
            \caption{Sample objects with their locations, orientations and dimensions.}
            \label{fig:hidden-objects}
        \end{figure}

        \paragraph{Train-test split}
        A key aspect of a dataset is how has been split into train and test sets. In particular, we have used $1075$ images for training and $269$ for testing. It is important to notice that the object present in the former sets will never be present in the latter in neither the same location nor orientation. Other than that, since the ``composed objects'', built mixing two ``basic shapes'', are randomly generated, the ones present in the training set will never be present in the test set. Regarding the front wall's roughness, $\alpha$ spans over the same interval in both sets ensuring that the model learns to handle different materials.

%% file: img/section_05/objects.tex
\subfloat{
    \includegraphics[width=.25\linewidth]{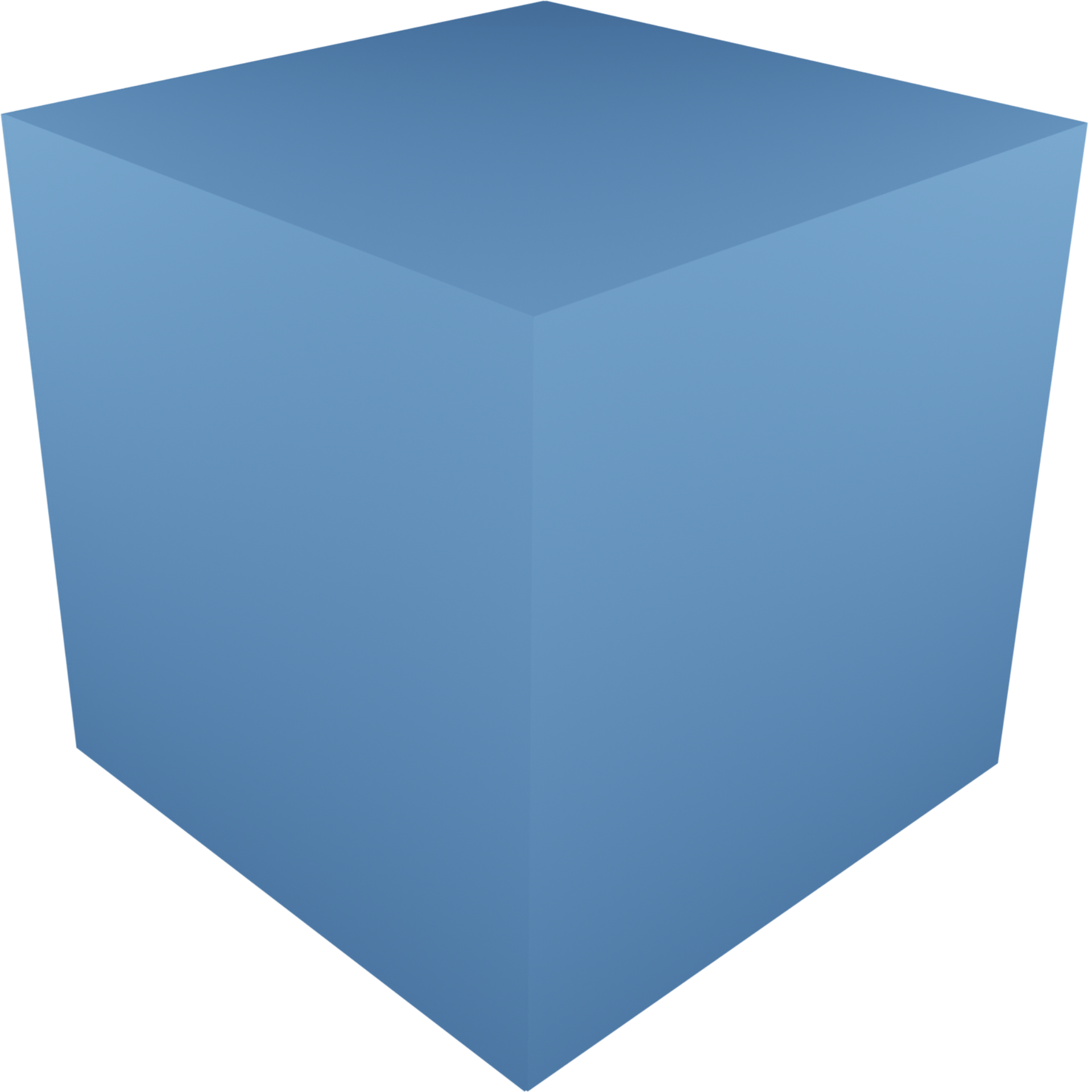}
    \label{subfig:hidden-objects:cube}
    }
\hfil
\subfloat{
    \includegraphics[width=.25\linewidth]{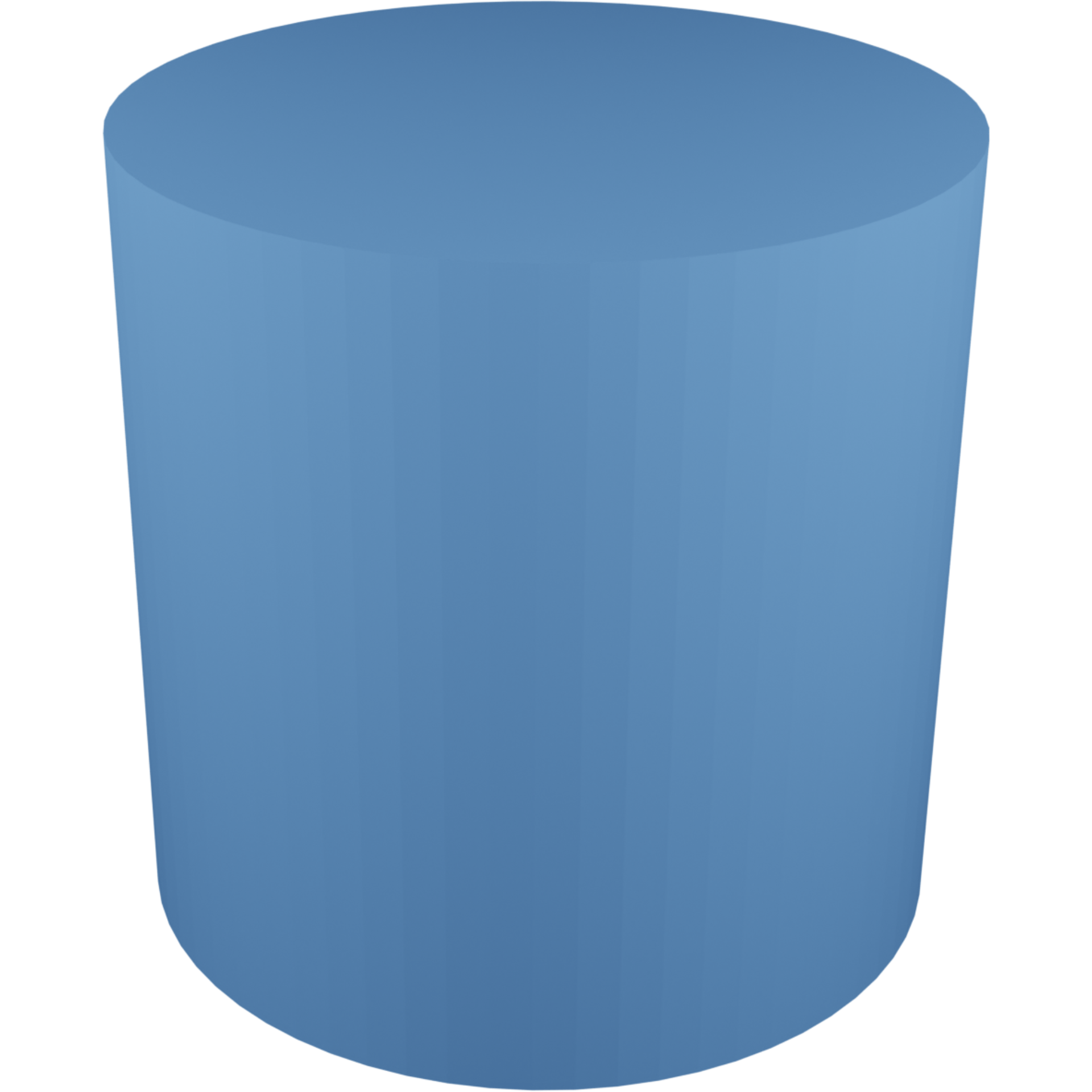}
    \label{subfig:hidden-objects:cylinder}
    }
\hfil
\subfloat{
    \includegraphics[width=.25\linewidth]{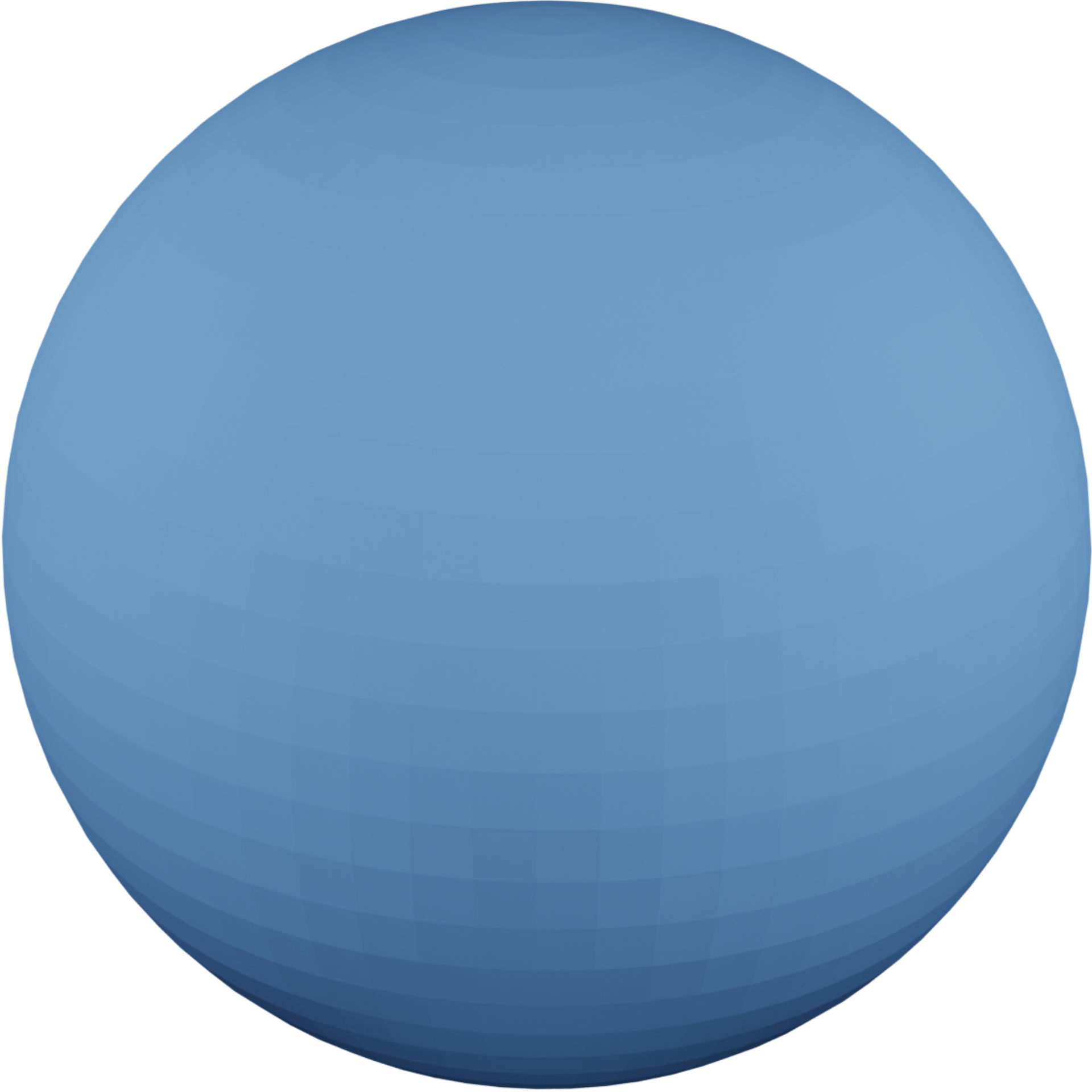}
    \label{subfig:hidden-objects:sphere}
    }
\\
\vspace{10pt}
\subfloat{
    \includegraphics[width=.25\linewidth]{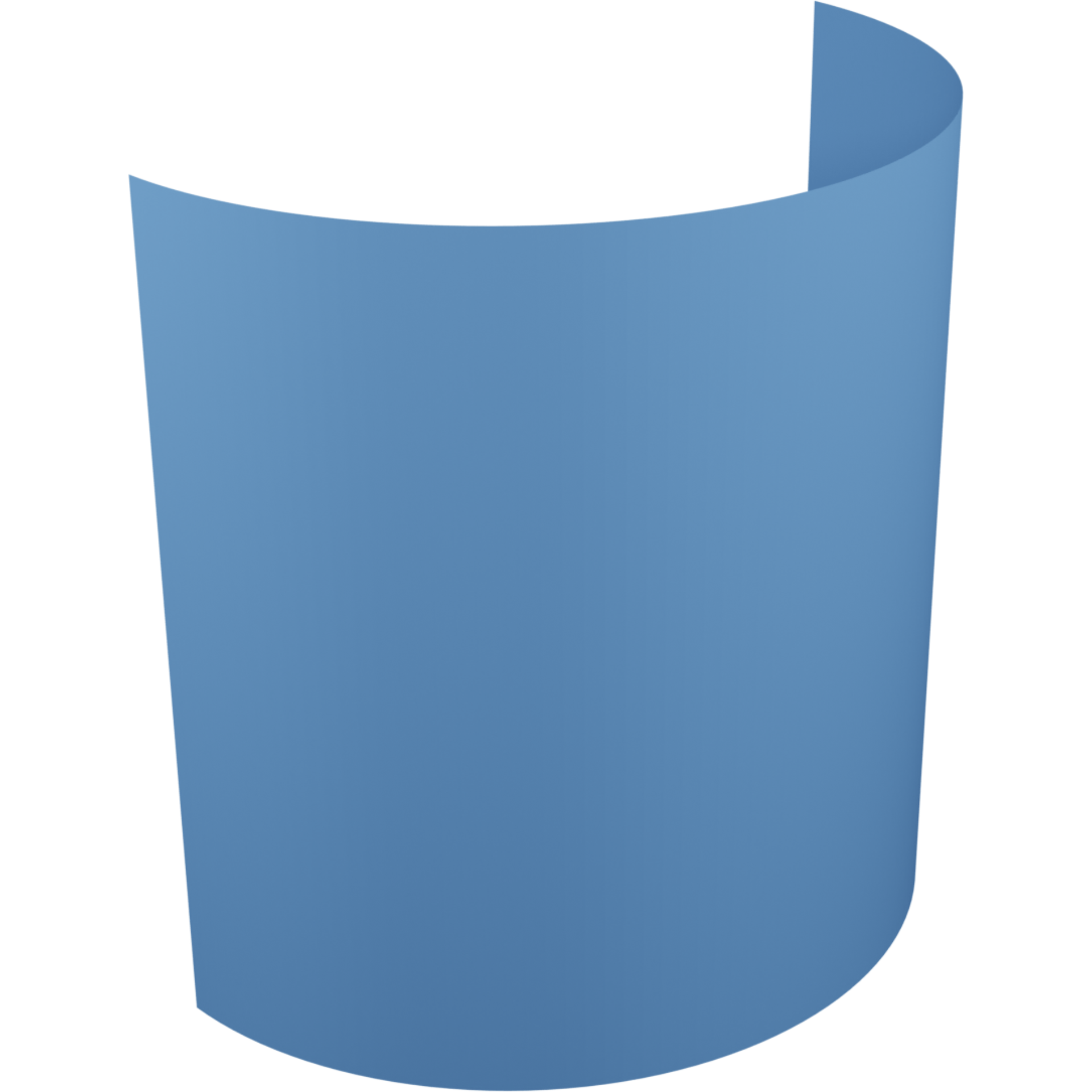}
    \label{subfig:hidden-objects:concave_plane}
    }
\hfil
\subfloat{
    \includegraphics[width=.25\linewidth]{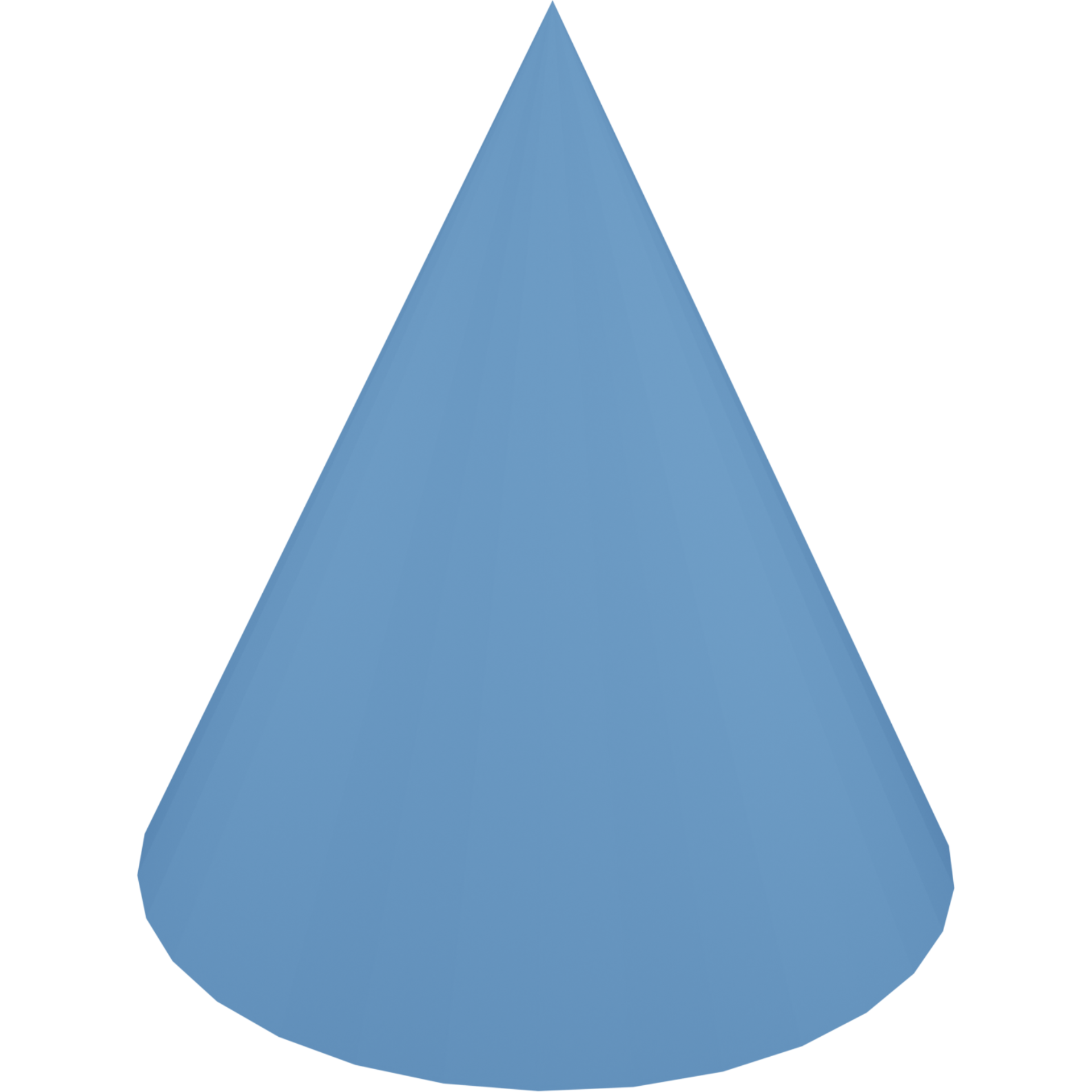}
    \label{subfig:hidden-objects:cone}
    }
\hfil
\subfloat{
    \includegraphics[width=.25\linewidth]{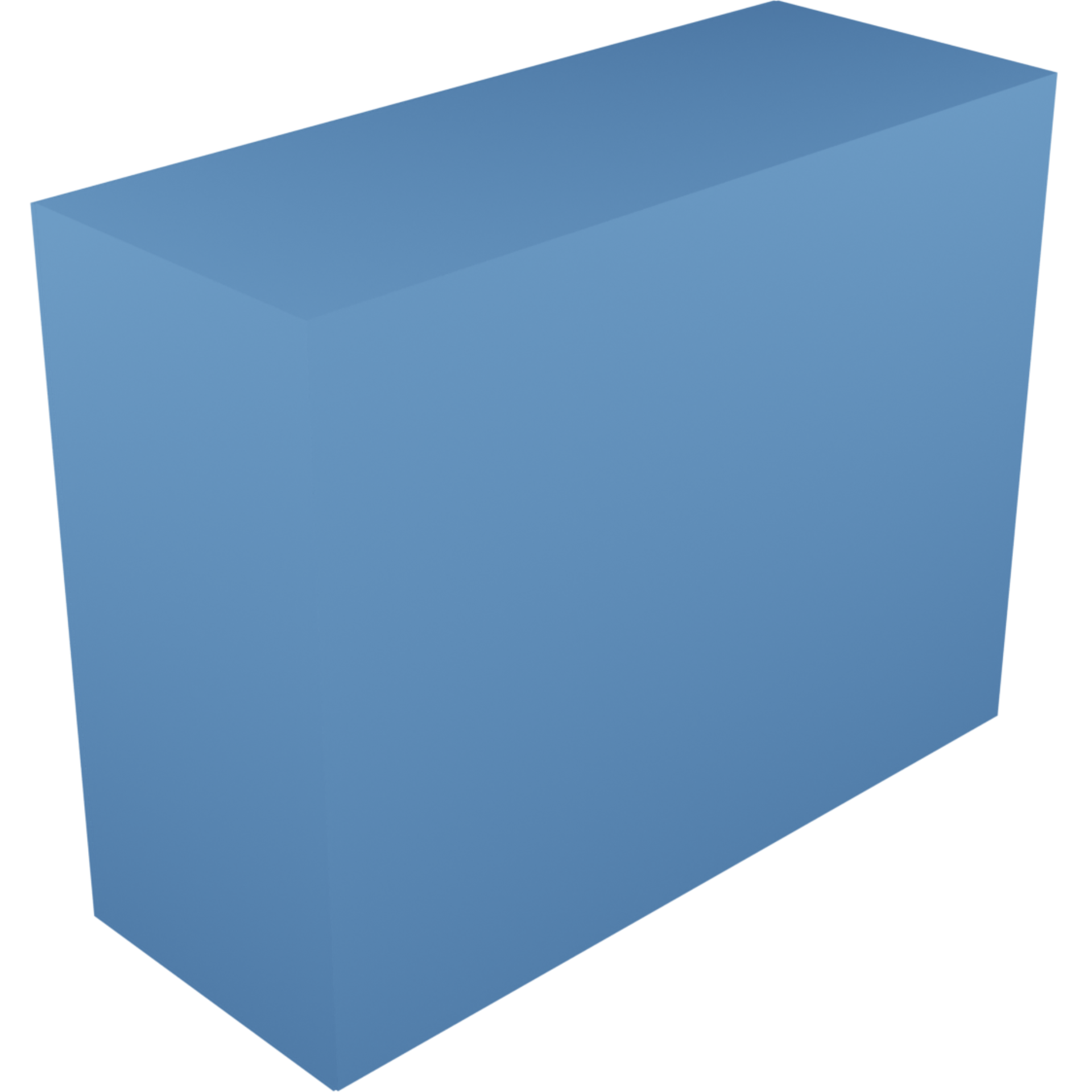}
    \label{subfig:hidden-objects:parallelepiped}
    }

%% file: tab/dataset_tab.tex
\begin{tabular}{cccc}\toprule
    \textbf{\textsc{object}}                                                          &    & \textbf{\textsc{position}} & \textbf{\textsc{rotation}} \\\midrule
    \multirow{3}{*}{\makecell{\textbf{front wall} \\ ($8 \times 4$)m}}                & X: & $0$m                       & $0^\circ$                  \\
                                                                                      & Y: & $0$m                       & $90^\circ$                 \\
                                                                                      & Z: & $2$m                       & $0^\circ$                  \\\midrule
    \multirow{3}{*}{\makecell{\textbf{middle wall} \\ ($3.30 \times 4$)m}}            & X: & $2.35$m                    & $90^\circ$                 \\
                                                                                      & Y: & $0$m                       & $0^\circ$                  \\
                                                                                      & Z: & $2$m                       & $0^\circ$                  \\\midrule
    \multirow{3}{*}{\textbf{source/detector}}                                         & X: & $1$m                       & $90^\circ$                 \\
                                                                                      & Y: & $-1$m                      & $0^\circ$                  \\
                                                                                      & Z: & $1.65$m                    & $50^\circ$                 \\\midrule
    \multirow{3}{*}{\makecell{\textbf{hidden object} \\ up to $0.5\textnormal{m}^3$}} & X: & $[1,\ 1.3]$m               & $[-90,\ 90]^\circ$         \\
                                                                                      & Y: & $[0.5,\ 1.3]$m             & $[-90,\ 90]^\circ$         \\
                                                                                      & Z: & $[1.25,\ 1.95]$m           & $[-90,\ 90]^\circ$         \\\bottomrule
\end{tabular}

%% file: sec/06_experiments.tex
\section{Experimental Results} \label{sec:experiments}
    We evaluated our approach over the test set of the introduced dataset. In order to properly assess the results we will consider both the accuracy of the reconstructed object's shapes and the depth estimation error. For the former, we computed the \ac{miou} between the predicted and \acl{gt} object shapes, obtained by flattening the depth map to a \twod image and considering as object everything that is not background. For evaluating the accuracy of the actual depth, instead, we will use the \acf{mae} between the estimated values and the \ac{gt} ones.
    In \cref{fig:results} we show some visual results. More precisely, we include five examples of predictions (one for each row). The first two columns contain the real ($\Re(c)$) and the imaginary component ($\Im(c)$) of the predicted \ac{itof} (output of the network), while the last column contains the depth map computed from the predictions. Each column compares the ground truth (left) with the corresponding prediction (right). From this image, it is possible to qualitatively see that the model can recover the overall shape of the target reasonably well, indeed our approach obtains an average \ac{miou} of $0.77$.
    Performing a more in-depth analysis we could see that the model performs best over cylindrical and spherical shapes, while the most challenging shape is the concave plane due to its thin shape, especially if not oriented straight to the camera. Regardless of that, it is important to note that from our tests the model is able to maintain almost the same reconstruction accuracy over simple objects and composed ones (\ie the scenes with two objects inside).
    
    \begin{table}[h]
        \centering
        \caption{Results over the test set. The plot shows the accuracy values distribution.}
        \begin{minipage}{0.49\textwidth}
            \centering
            \resizebox{.9\textwidth}{!}{%
            \input{tab/results_score}}
        \end{minipage}
        \begin{minipage}{0.49\textwidth}
            \centering
            \resizebox{.65\textwidth}{!}{%
            \input{img/section_06/losses_histograms}}
        \end{minipage}
        \label{tab:results_scores}
    \end{table}
    
    \noindent For a more quantitative analysis, we can refer to the plot in \cref{tab:results_scores}: most of the sample predictions reach an overall accuracy, over the test set, between $80\%$ and $95\%$. This represents a good result considering the very challenging scenario in which we are working. 
    Indeed, considering that we illuminate the scene \textit{full-field} reducing the acquisition to a single shoot, differently from the long acquisitions of most \acs{dtof}-based approaches, the achieved performance is really promising. 
    In \cref{tab:results_scores} we also present the depth estimation error. It is possible to see that the network recovers the depth of the scene with an average error of $5.21$cm, which, considering that the mean distance between the sensor and the object is around $3$m is an encouraging result (the error is around $1-2\%$ of the distance). 
    It is important to note that in some cases the model can make an almost perfect prediction (in the best case, the error is only $\approx 0.11$cm).
    To better visualize the depth reconstruction of the model, in \cref{fig:pc} we also show the point cloud of each predicted sample of \cref{fig:results} together with the related ground truth. Analyzing \cref{fig:results,fig:pc} together we can confirm the evaluation done on the quantitative results. We can see that the model is able to correctly locate the hidden object and predict its dimensions. The most challenging aspect is the precise reconstruction of the shape. The first two samples have been recovered almost perfectly, while the last two shapes are not so accurate. Considering the third example, we notice that the network struggles a bit with the complex shape of the composed object, but is still able to predict a rough shape that, rather not being as accurate as the one of the cylinder or sphere, is still quite good.
    \input{img/section_06/results}

    \input{img/section_06/pc}

    \subsection{Ablation study} \label{sub:experiments:ablation}
        Since there is no other work suitable for tackling the task for direct comparison, it is very interesting to show how the various design choices allow us to improve results \wrt baseline strategies. The findings of this analysis are in \cref{tab:ablation_scores}.
        
        \begin{table}[h]
            \caption{Ablation scores for object shape (\ac{miou}) and depth (\ac{mae}) over the test set.}
            \label{tab:ablation_scores}
            \centering
            \resizebox{.75\textwidth}{!}{
            \begin{tabular}{l@{\hskip 15pt} c@{\hskip 12pt} c c c@{\hskip 12pt} c@{\hskip 12pt} c@{\hskip 15pt} c@{\hskip 15pt}}
                \multirow{2}{*}{\textsc{settings}} & \multicolumn{3}{c@{\hskip 15pt}}{frequencies (MHz)} & \multicolumn{3}{c@{\hskip 15pt}}{losses}      & \multirow{2}{*}{\makecell{\textbf{our model}\\\textbf{\acs{mae} + \acs{iou}}}} \\\cmidrule(l{-3pt}r{10pt}){2-4} \cmidrule(r{12pt}){5-7}
                                                   & $20$   & $50$    & $60$                             & \acs{mae} & \acs{mse} & \acs{mse} + \acs{iou} &                            \\\toprule
                \acs{mae} [cm] $\downarrow$        & $6.15$ & $10.33$ & $9.43$                           & $5.83$    & $13.72$   & $5.93$                & $\bm{5.21}$                \\
                \acs{miou} $\uparrow$              & $0.74$ & $0.68$  & $0.69$                           & $0.76$    & $0.66$    & $0.77$                & $\bm{0.77}$                \\\bottomrule
            \end{tabular}}
        \end{table}

        \paragraph{Input Frequencies}
        Using multi-frequency data greatly helps the \ac{dnn} to reach higher accuracy. Due to how an \ac{itof} works, higher frequencies ensure higher resolution but suffer from the \textit{phase wrapping} \cite{zanuttigh2016time}  issue. On the other hand, using only the $20$MHz frequency, we greatly reduce the influence of the \textit{phase wrapping} but also affect the accuracy of the reconstruction. Taking into account this evaluation, it is clear that merging acquisitions at different frequencies leads to exploiting the strengths of each of them, minimizing the drawbacks.

        \paragraph{Loss Functions}
        Training the model using only the \ac{mae} loss, the overall accuracy achieved by the network decreases, proving that the \ac{iou} metric helps to better identify the object. Furthermore, if we consider the qualitative results obtained by removing the \ac{iou} it is clear that the shape of the object is completely lost since the model tends to build a spherical object. If we use the \ac{mse} loss instead of the \ac{mae} we can see that the error greatly increases, probably because more focus on smaller errors is needed. Adding \ac{iou} to \ac{mse} improves the model accuracy, but \ac{mae} with \ac{iou} remains the optimal.

%% file: tab/results_score.tex
\setlength{\tabcolsep}{8pt}
\begin{tabular}{l l l l}
                                & \textsc{average}      & \textsc{min}  & \textsc{max} \\\toprule
    \acs{mae} [cm] $\downarrow$ & $\bm{5.21} \pm 2.84$  & $1.14$        & $18.70$      \\
    \acs{miou} $\uparrow$       & $\bm{0.77} \pm 0.12$  & $0.48$        & $0.95$       \\\bottomrule
\end{tabular}

%% file: img/section_06/losses_histograms.tex
\begin{tikzpicture}
    \definecolor{darkslategray38}{RGB}{38,38,38}
    \definecolor{lavender234234242}{RGB}{234,234,242}
    \definecolor{steelblue76114176}{RGB}{76,114,176}

    \begin{groupplot}[group style={group size=1 by 1}]
        \nextgroupplot[
            width = 12cm, 
            height = 7.5cm,
            axis background/.style={fill=lavender234234242},
            axis line style={white},
            tick align=outside,
            x grid style={white},
            xlabel=\textcolor{darkslategray38}{Accuracy value},
            xmajorgrids,
            xmajorticks=true,
            xmin=0.656015901034698, xmax=0.982480025710538,
            xtick style={color=darkslategray38, draw=none},
            y grid style={white},
            ylabel=\textcolor{darkslategray38}{Number of occurrences},
            ymajorgrids,
            ymajorticks=true,
            ymin=0, ymax=21,
            ytick style={color=darkslategray38, draw=none}
        ]

        \draw[draw=white,fill=steelblue76114176] (axis cs:0.670855179429054,0) rectangle (axis cs:0.690640883954863,2);
        \draw[draw=white,fill=steelblue76114176] (axis cs:0.690640883954863,0) rectangle (axis cs:0.710426588480671,6);
        \draw[draw=white,fill=steelblue76114176] (axis cs:0.710426588480671,0) rectangle (axis cs:0.73021229300648,2);
        \draw[draw=white,fill=steelblue76114176] (axis cs:0.73021229300648,0) rectangle (axis cs:0.749997997532288,5);
        \draw[draw=white,fill=steelblue76114176] (axis cs:0.749997997532288,0) rectangle (axis cs:0.769783702058097,2);
        \draw[draw=white,fill=steelblue76114176] (axis cs:0.769783702058097,0) rectangle (axis cs:0.789569406583905,2);
        \draw[draw=white,fill=steelblue76114176] (axis cs:0.789569406583905,0) rectangle (axis cs:0.809355111109714,8);
        \draw[draw=white,fill=steelblue76114176] (axis cs:0.809355111109714,0) rectangle (axis cs:0.829140815635522,8);
        \draw[draw=white,fill=steelblue76114176] (axis cs:0.829140815635522,0) rectangle (axis cs:0.848926520161331,14);
        \draw[draw=white,fill=steelblue76114176] (axis cs:0.848926520161331,0) rectangle (axis cs:0.868712224687139,15);
        \draw[draw=white,fill=steelblue76114176] (axis cs:0.868712224687139,0) rectangle (axis cs:0.888497929212948,20);
        \draw[draw=white,fill=steelblue76114176] (axis cs:0.888497929212948,0) rectangle (axis cs:0.908283633738756,17);
        \draw[draw=white,fill=steelblue76114176] (axis cs:0.908283633738756,0) rectangle (axis cs:0.928069338264565,12);
        \draw[draw=white,fill=steelblue76114176] (axis cs:0.928069338264565,0) rectangle (axis cs:0.947855042790373,12);
        \draw[draw=white,fill=steelblue76114176] (axis cs:0.947855042790373,0) rectangle (axis cs:0.967640747316182,9);

        \draw ({$(current bounding box.south west)!0.5!(current bounding box.south east)$}|-{$(current bounding box.south west)!0.98!(current bounding box.north west)$}) node[
          scale=0.7,
          anchor=north,
          text=darkslategray38,
          rotate=0.0
        ]{};
    \end{groupplot}
\end{tikzpicture}

%% file: img/section_06/results.tex
\ActivateWarningFilters[pgflabel]
    \begin{figure}[!tb]
        \centering
        \subfloat{
            \resizebox{.98\linewidth}{!}{
            \input{img/section_06/prediction_01/2d_plots/25}
            }
        }
        \vspace{0.3em}
        \subfloat{
            \resizebox{.98\linewidth}{!}{
            \input{img/section_06/prediction_02/2d_plots/62}
            }
        }
        \vspace{0.3em}
        \subfloat{
            \resizebox{.98\linewidth}{!}{
            \input{img/section_06/prediction_03/2d_plots/28}
            }
        }
        \vspace{0.3em}
        \subfloat{
            \resizebox{.98\linewidth}{!}{
            \input{img/section_06/prediction_04/2d_plots/74}
            }
        }
        \vspace{0.3em}
        \subfloat{
            \resizebox{.98\linewidth}{!}{
            \input{img/section_06/prediction_05/2d_plots/31}
            }
        }
        \caption{Sample visual results}
        \label{fig:results}
    \end{figure}
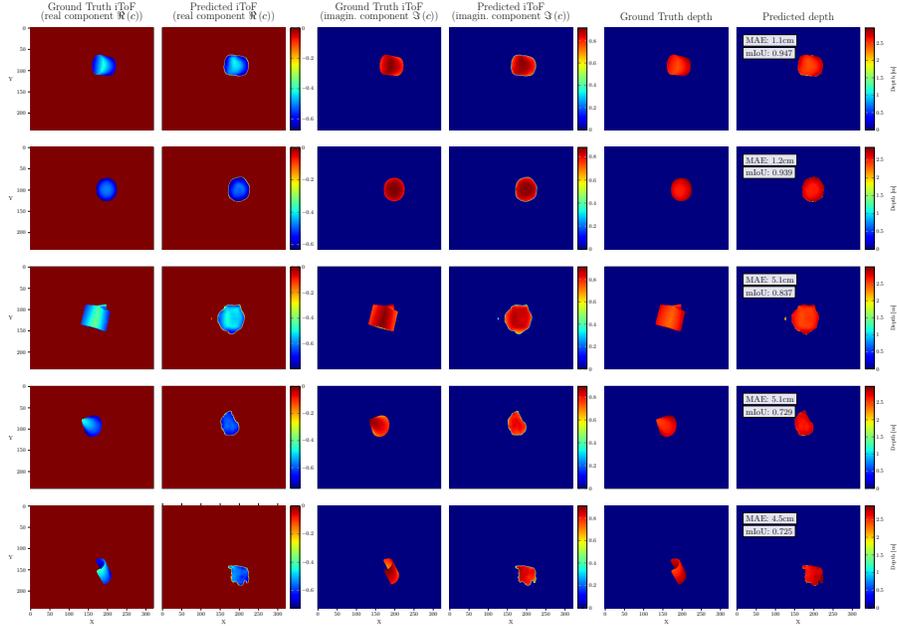
\DeactivateWarningFilters[pgflabel]

%% file: img/section_06/prediction_01/2d_plots/25.tex
\newcommand{\imgfolder}{img/section_06/prediction_01/2d_plots/}

\begin{tikzpicture}
    \definecolor{darkgray176}{RGB}{176,176,176}

    \begin{groupplot}[group style={group size=8 by 1,
                                   vertical sep=1.9cm,
                                   horizontal sep=0.5cm}]
        \nextgroupplot[
            colormap={mymap}{[1pt]
                             rgb(0pt)=(0,0,0.5);
                             rgb(22pt)=(0,0,1);
                             rgb(25pt)=(0,0,1);
                             rgb(68pt)=(0,0.86,1);
                             rgb(70pt)=(0,0.9,0.967741935483871);
                             rgb(75pt)=(0.0806451612903226,1,0.887096774193548);
                             rgb(128pt)=(0.935483870967742,1,0.0322580645161291);
                             rgb(130pt)=(0.967741935483871,0.962962962962963,0);
                             rgb(132pt)=(1,0.925925925925926,0);
                             rgb(178pt)=(1,0.0740740740740741,0);
                             rgb(182pt)=(0.909090909090909,0,0);
                             rgb(200pt)=(0.5,0,0)
            },
            point meta max=0,
            point meta min=-0.672179222106934,
            tick align=outside,
            align=center,
            title={\LARGE{Ground Truth iToF} \\ \LARGE{(real component $\Re \left(c\right)$)}},
            x grid style={darkgray176},
            xlabel={},
            xmajorticks=false,
            xmin=-0.5, xmax=319.5,
            xtick pos=bottom,
            xtick style={color=black},
            y dir=reverse,
            y grid style={darkgray176},
            ylabel style={rotate=-90.0},
            ylabel={Y},
            ymajorticks=true,
            ymin=-0.5, ymax=239.5,
            ytick pos=left,
            ytick style={color=black}
            ]
            \addplot graphics [includegraphics cmd=\pgfimage,xmin=-0.5, xmax=319.5, ymin=239.5, ymax=-0.5] {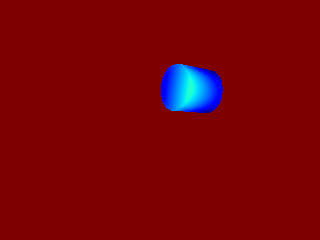};
        
        \nextgroupplot[
        colorbar,
        colorbar style={
            ylabel style={
                yshift=-9em
            }
        },
        colormap={mymap}{[1pt]
          rgb(0pt)=(0,0,0.5);
          rgb(22pt)=(0,0,1);
          rgb(25pt)=(0,0,1);
          rgb(68pt)=(0,0.86,1);
          rgb(70pt)=(0,0.9,0.967741935483871);
          rgb(75pt)=(0.0806451612903226,1,0.887096774193548);
          rgb(128pt)=(0.935483870967742,1,0.0322580645161291);
          rgb(130pt)=(0.967741935483871,0.962962962962963,0);
          rgb(132pt)=(1,0.925925925925926,0);
          rgb(178pt)=(1,0.0740740740740741,0);
          rgb(182pt)=(0.909090909090909,0,0);
          rgb(200pt)=(0.5,0,0)
        },
        point meta max=0,
        point meta min=-0.672179222106934,
        tick align=outside,
        align=center,
        title={\LARGE{Predicted iToF} \\ \LARGE{(real component $\Re \left(c\right)$)}},
        x grid style={darkgray176},
        xlabel={},
        xmajorticks=false,
        xmin=-0.5, xmax=319.5,
        xtick pos=bottom,
        xtick style={color=black},
        y dir=reverse,
        y grid style={darkgray176},
        ylabel style={rotate=-90.0},
        ylabel={},
        ymajorticks=false,
        ymin=-0.5, ymax=239.5,
        ytick pos=left,
        ytick style={color=black}
        ]
        \addplot graphics [includegraphics cmd=\pgfimage,xmin=-0.5, xmax=319.5, ymin=239.5, ymax=-0.5] {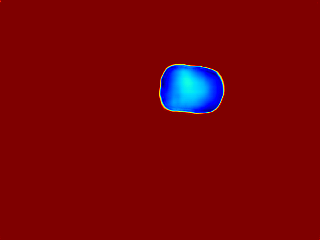};
        
        \nextgroupplot[group/empty plot, width=2.4cm] %
        
        \nextgroupplot[
        colormap={mymap}{[1pt]
          rgb(0pt)=(0,0,0.5);
          rgb(22pt)=(0,0,1);
          rgb(25pt)=(0,0,1);
          rgb(68pt)=(0,0.86,1);
          rgb(70pt)=(0,0.9,0.967741935483871);
          rgb(75pt)=(0.0806451612903226,1,0.887096774193548);
          rgb(128pt)=(0.935483870967742,1,0.0322580645161291);
          rgb(130pt)=(0.967741935483871,0.962962962962963,0);
          rgb(132pt)=(1,0.925925925925926,0);
          rgb(178pt)=(1,0.0740740740740741,0);
          rgb(182pt)=(0.909090909090909,0,0);
          rgb(200pt)=(0.5,0,0)
        },
        point meta max=0.91536283493042,
        point meta min=0,
        tick align=outside,
        align=center,
        title={\LARGE{Ground Truth iToF} \\ \LARGE{(imagin. component $\Im \left(c\right)$)}},
        x grid style={darkgray176},
        xlabel={},
        xmajorticks=false,
        xmin=-0.5, xmax=319.5,
        xtick pos=bottom,
        xtick style={color=black},
        y dir=reverse,
        y grid style={darkgray176},
        ylabel style={rotate=-90.0},
        ylabel={},
        ymajorticks=false,
        ymin=-0.5, ymax=239.5,
        ytick pos=left,
        ytick style={color=black}
        ]
        \addplot graphics [includegraphics cmd=\pgfimage,xmin=-0.5, xmax=319.5, ymin=239.5, ymax=-0.5] {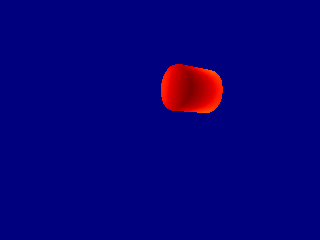};
        
        \nextgroupplot[
        colorbar,
        colorbar style={
            ylabel style={
                yshift=-9em
            }
        },
        colormap={mymap}{[1pt]
          rgb(0pt)=(0,0,0.5);
          rgb(22pt)=(0,0,1);
          rgb(25pt)=(0,0,1);
          rgb(68pt)=(0,0.86,1);
          rgb(70pt)=(0,0.9,0.967741935483871);
          rgb(75pt)=(0.0806451612903226,1,0.887096774193548);
          rgb(128pt)=(0.935483870967742,1,0.0322580645161291);
          rgb(130pt)=(0.967741935483871,0.962962962962963,0);
          rgb(132pt)=(1,0.925925925925926,0);
          rgb(178pt)=(1,0.0740740740740741,0);
          rgb(182pt)=(0.909090909090909,0,0);
          rgb(200pt)=(0.5,0,0)
        },
        point meta max=0.91536283493042,
        point meta min=0,
        tick align=outside,
        align=center,
        title={\LARGE{Predicted iToF} \\ \LARGE{(imagin. component $\Im \left(c\right)$)}},
        x grid style={darkgray176},
        xlabel={},
        xmajorticks=false,
        xmin=-0.5, xmax=319.5,
        xtick pos=bottom,
        xtick style={color=black},
        y dir=reverse,
        y grid style={darkgray176},
        ylabel style={rotate=-90.0},
        ylabel={},
        ymajorticks=false,
        ymin=-0.5, ymax=239.5,
        ytick pos=left,
        ytick style={color=black}
        ]
        \addplot graphics [includegraphics cmd=\pgfimage,xmin=-0.5, xmax=319.5, ymin=239.5, ymax=-0.5] {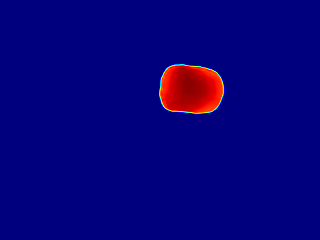};
        
        \nextgroupplot[group/empty plot, width=2.4cm] %
        
        \nextgroupplot[
        colormap={mymap}{[1pt]
          rgb(0pt)=(0,0,0.5);
          rgb(22pt)=(0,0,1);
          rgb(25pt)=(0,0,1);
          rgb(68pt)=(0,0.86,1);
          rgb(70pt)=(0,0.9,0.967741935483871);
          rgb(75pt)=(0.0806451612903226,1,0.887096774193548);
          rgb(128pt)=(0.935483870967742,1,0.0322580645161291);
          rgb(130pt)=(0.967741935483871,0.962962962962963,0);
          rgb(132pt)=(1,0.925925925925926,0);
          rgb(178pt)=(1,0.0740740740740741,0);
          rgb(182pt)=(0.909090909090909,0,0);
          rgb(200pt)=(0.5,0,0)
        },
        point meta max=2.93590609302592,
        point meta min=0,
        tick align=outside,
        title={\LARGE{Ground Truth depth}},
        x grid style={darkgray176},
        xlabel={},
        xmajorticks=false,
        xmin=-0.5, xmax=319.5,
        xtick pos=bottom,
        xtick style={color=black},
        y dir=reverse,
        y grid style={darkgray176},
        ylabel style={rotate=-90.0},
        ylabel={},
        ymajorticks=false,
        ymin=-0.5, ymax=239.5,
        ytick pos=left,
        ytick style={color=black}
        ]
        \addplot graphics [includegraphics cmd=\pgfimage,xmin=-0.5, xmax=319.5, ymin=239.5, ymax=-0.5] {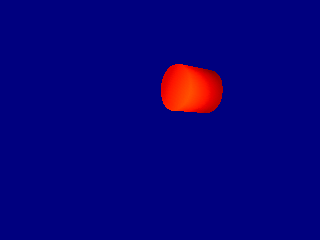};
        
        \nextgroupplot[
        colorbar,
        colorbar style={
            ylabel={Depth [m]},
            ylabel style={
                yshift=-.5em
            }
        },
        colormap={mymap}{[1pt]
          rgb(0pt)=(0,0,0.5);
          rgb(22pt)=(0,0,1);
          rgb(25pt)=(0,0,1);
          rgb(68pt)=(0,0.86,1);
          rgb(70pt)=(0,0.9,0.967741935483871);
          rgb(75pt)=(0.0806451612903226,1,0.887096774193548);
          rgb(128pt)=(0.935483870967742,1,0.0322580645161291);
          rgb(130pt)=(0.967741935483871,0.962962962962963,0);
          rgb(132pt)=(1,0.925925925925926,0);
          rgb(178pt)=(1,0.0740740740740741,0);
          rgb(182pt)=(0.909090909090909,0,0);
          rgb(200pt)=(0.5,0,0)
        },
        point meta max=2.93590609302592,
        point meta min=0,
        tick align=outside,
        title={\LARGE{Predicted depth}},
        x grid style={darkgray176},
        xlabel={},
        xmajorticks=false,
        xmin=-0.5, xmax=319.5,
        xtick pos=bottom,
        xtick style={color=black},
        y dir=reverse,
        y grid style={darkgray176},
        ylabel style={rotate=-90.0},
        ylabel={},
        ymajorticks=false,
        ymin=-0.5, ymax=239.5,
        ytick pos=left,
        ytick style={color=black}
        ]
        \addplot graphics [includegraphics cmd=\pgfimage,xmin=-0.5, xmax=319.5, ymin=239.5, ymax=-0.5] {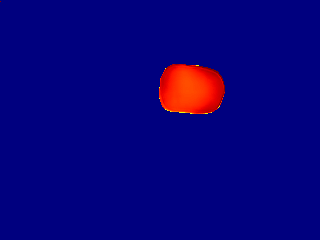};
        \draw (axis cs:15,15) node[
          scale=1.5,
          fill=white,
          draw=black,
          line width=0.4pt,
          inner sep=3pt,
          fill opacity=0.9,
          rounded corners=0.5mm,
          anchor=north west,
          text=black,
          rotate=0.0
        ]{MAE: $1.1$cm};
        \draw (axis cs:15,45) node[
          scale=1.5,
          fill=white,
          draw=black,
          line width=0.4pt,
          inner sep=3pt,
          fill opacity=0.9,
          rounded corners=0.5mm,
          anchor=north west,
          text=black,
          rotate=0.0
        ]{mIoU: $0.947$};

\end{groupplot}

\end{tikzpicture}

%% file: img/section_06/prediction_02/2d_plots/62.tex
\newcommand{\imgfolder}{img/section_06/prediction_02/2d_plots/}

\begin{tikzpicture}

\definecolor{darkgray176}{RGB}{176,176,176}

\begin{groupplot}[
    group style={
        group size=8 by 1,
        vertical sep=1.9cm,
        horizontal sep=0.5cm
    }
]

\nextgroupplot[
colormap={mymap}{[1pt]
  rgb(0pt)=(0,0,0.5);
  rgb(22pt)=(0,0,1);
  rgb(25pt)=(0,0,1);
  rgb(68pt)=(0,0.86,1);
  rgb(70pt)=(0,0.9,0.967741935483871);
  rgb(75pt)=(0.0806451612903226,1,0.887096774193548);
  rgb(128pt)=(0.935483870967742,1,0.0322580645161291);
  rgb(130pt)=(0.967741935483871,0.962962962962963,0);
  rgb(132pt)=(1,0.925925925925926,0);
  rgb(178pt)=(1,0.0740740740740741,0);
  rgb(182pt)=(0.909090909090909,0,0);
  rgb(200pt)=(0.5,0,0)
},
point meta max=0,
point meta min=-0.629554510116577,
tick align=outside,
x grid style={darkgray176},
xlabel={},
xmajorticks=false,
xmin=-0.5, xmax=319.5,
xtick pos=both,
xtick style={color=black},
y dir=reverse,
y grid style={darkgray176},
ylabel style={rotate=-90.0},
ylabel={Y},
ymin=-0.5, ymax=239.5,
ytick pos=left,
ytick style={color=black}
]
\addplot graphics [includegraphics cmd=\pgfimage,xmin=-0.5, xmax=319.5, ymin=239.5, ymax=-0.5] {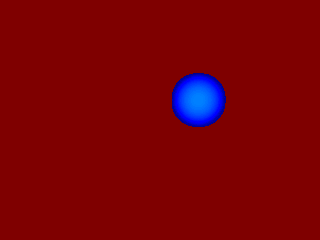};

\nextgroupplot[
colorbar,
colormap={mymap}{[1pt]
  rgb(0pt)=(0,0,0.5);
  rgb(22pt)=(0,0,1);
  rgb(25pt)=(0,0,1);
  rgb(68pt)=(0,0.86,1);
  rgb(70pt)=(0,0.9,0.967741935483871);
  rgb(75pt)=(0.0806451612903226,1,0.887096774193548);
  rgb(128pt)=(0.935483870967742,1,0.0322580645161291);
  rgb(130pt)=(0.967741935483871,0.962962962962963,0);
  rgb(132pt)=(1,0.925925925925926,0);
  rgb(178pt)=(1,0.0740740740740741,0);
  rgb(182pt)=(0.909090909090909,0,0);
  rgb(200pt)=(0.5,0,0)
},
point meta max=0,
point meta min=-0.629554510116577,
tick align=outside,
x grid style={darkgray176},
xlabel={},
xmajorticks=false,
xmin=-0.5, xmax=319.5,
xtick pos=both,
xtick style={color=black},
y dir=reverse,
y grid style={darkgray176},
ylabel style={rotate=-90.0},
ylabel={},
ymajorticks=false,
ymin=-0.5, ymax=239.5,
ytick pos=left,
ytick style={color=black}
]
\addplot graphics [includegraphics cmd=\pgfimage,xmin=-0.5, xmax=319.5, ymin=239.5, ymax=-0.5] {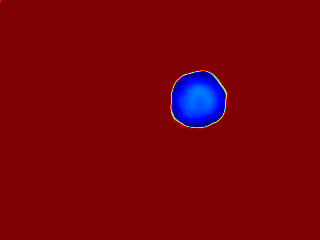};

\nextgroupplot[group/empty plot, width=2.4cm] %

\nextgroupplot[
colormap={mymap}{[1pt]
  rgb(0pt)=(0,0,0.5);
  rgb(22pt)=(0,0,1);
  rgb(25pt)=(0,0,1);
  rgb(68pt)=(0,0.86,1);
  rgb(70pt)=(0,0.9,0.967741935483871);
  rgb(75pt)=(0.0806451612903226,1,0.887096774193548);
  rgb(128pt)=(0.935483870967742,1,0.0322580645161291);
  rgb(130pt)=(0.967741935483871,0.962962962962963,0);
  rgb(132pt)=(1,0.925925925925926,0);
  rgb(178pt)=(1,0.0740740740740741,0);
  rgb(182pt)=(0.909090909090909,0,0);
  rgb(200pt)=(0.5,0,0)
},
point meta max=0.882408857345581,
point meta min=0,
tick align=outside,
x grid style={darkgray176},
xlabel={},
xmajorticks=false,
xmin=-0.5, xmax=319.5,
xtick pos=both,
xtick style={color=black},
y dir=reverse,
y grid style={darkgray176},
ylabel style={rotate=-90.0},
ylabel={},
ymajorticks=false,
ymin=-0.5, ymax=239.5,
ytick pos=left,
ytick style={color=black}
]
\addplot graphics [includegraphics cmd=\pgfimage,xmin=-0.5, xmax=319.5, ymin=239.5, ymax=-0.5] {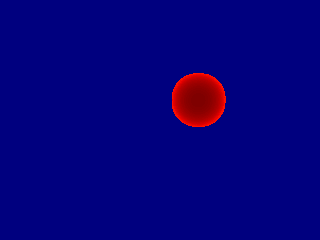};

\nextgroupplot[
colorbar,
colormap={mymap}{[1pt]
  rgb(0pt)=(0,0,0.5);
  rgb(22pt)=(0,0,1);
  rgb(25pt)=(0,0,1);
  rgb(68pt)=(0,0.86,1);
  rgb(70pt)=(0,0.9,0.967741935483871);
  rgb(75pt)=(0.0806451612903226,1,0.887096774193548);
  rgb(128pt)=(0.935483870967742,1,0.0322580645161291);
  rgb(130pt)=(0.967741935483871,0.962962962962963,0);
  rgb(132pt)=(1,0.925925925925926,0);
  rgb(178pt)=(1,0.0740740740740741,0);
  rgb(182pt)=(0.909090909090909,0,0);
  rgb(200pt)=(0.5,0,0)
},
point meta max=0.882408857345581,
point meta min=0,
tick align=outside,
x grid style={darkgray176},
xlabel={},
xmajorticks=false,
xmin=-0.5, xmax=319.5,
xtick pos=both,
xtick style={color=black},
y dir=reverse,
y grid style={darkgray176},
ylabel style={rotate=-90.0},
ylabel={},
ymajorticks=false,
ymin=-0.5, ymax=239.5,
ytick pos=left,
ytick style={color=black}
]
\addplot graphics [includegraphics cmd=\pgfimage,xmin=-0.5, xmax=319.5, ymin=239.5, ymax=-0.5] {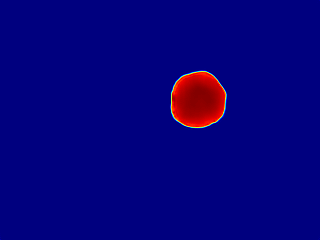};

\nextgroupplot[group/empty plot, width=2.4cm] %

\nextgroupplot[
colormap={mymap}{[1pt]
  rgb(0pt)=(0,0,0.5);
  rgb(22pt)=(0,0,1);
  rgb(25pt)=(0,0,1);
  rgb(68pt)=(0,0.86,1);
  rgb(70pt)=(0,0.9,0.967741935483871);
  rgb(75pt)=(0.0806451612903226,1,0.887096774193548);
  rgb(128pt)=(0.935483870967742,1,0.0322580645161291);
  rgb(130pt)=(0.967741935483871,0.962962962962963,0);
  rgb(132pt)=(1,0.925925925925926,0);
  rgb(178pt)=(1,0.0740740740740741,0);
  rgb(182pt)=(0.909090909090909,0,0);
  rgb(200pt)=(0.5,0,0)
},
point meta max=2.85276534709657,
point meta min=0,
tick align=outside,
x grid style={darkgray176},
xlabel={},
xmajorticks=false,
xmin=-0.5, xmax=319.5,
xtick pos=both,
xtick style={color=black},
y dir=reverse,
y grid style={darkgray176},
ylabel style={rotate=-90.0},
ylabel={},
ymajorticks=false,
ymin=-0.5, ymax=239.5,
ytick pos=left,
ytick style={color=black}
]
\addplot graphics [includegraphics cmd=\pgfimage,xmin=-0.5, xmax=319.5, ymin=239.5, ymax=-0.5] {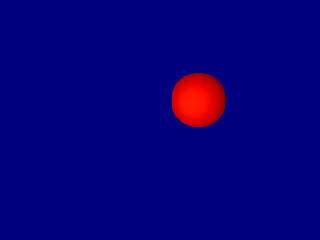};

\nextgroupplot[
colorbar,
colorbar style={
    ylabel={Depth [m]},
    ylabel style={
        yshift=-.5em
    }
},
colormap={mymap}{[1pt]
  rgb(0pt)=(0,0,0.5);
  rgb(22pt)=(0,0,1);
  rgb(25pt)=(0,0,1);
  rgb(68pt)=(0,0.86,1);
  rgb(70pt)=(0,0.9,0.967741935483871);
  rgb(75pt)=(0.0806451612903226,1,0.887096774193548);
  rgb(128pt)=(0.935483870967742,1,0.0322580645161291);
  rgb(130pt)=(0.967741935483871,0.962962962962963,0);
  rgb(132pt)=(1,0.925925925925926,0);
  rgb(178pt)=(1,0.0740740740740741,0);
  rgb(182pt)=(0.909090909090909,0,0);
  rgb(200pt)=(0.5,0,0)
},
point meta max=2.85276534709657,
point meta min=0,
tick align=outside,
x grid style={darkgray176},
xlabel={},
xmajorticks=false,
xmin=-0.5, xmax=319.5,
xtick pos=both,
xtick style={color=black},
y dir=reverse,
y grid style={darkgray176},
ylabel style={rotate=-90.0},
ylabel={},
ymajorticks=false,
ymin=-0.5, ymax=239.5,
ytick pos=left,
ytick style={color=black}
]
\addplot graphics [includegraphics cmd=\pgfimage,xmin=-0.5, xmax=319.5, ymin=239.5, ymax=-0.5] {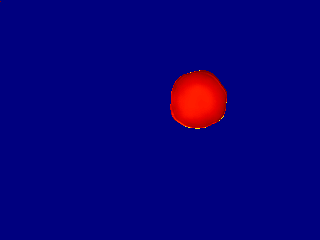};
\draw (axis cs:15,15) node[
  scale=1.5,
  fill=white,
  draw=black,
  line width=0.4pt,
  inner sep=3pt,
  fill opacity=0.9,
  rounded corners=0.5mm,
  anchor=north west,
  text=black,
  rotate=0.0
]{MAE: $1.2$cm};
\draw (axis cs:15,45) node[
  scale=1.5,
  fill=white,
  draw=black,
  line width=0.4pt,
  inner sep=3pt,
  fill opacity=0.9,
  rounded corners=0.5mm,
  anchor=north west,
  text=black,
  rotate=0.0
]{mIoU: $0.939$};

\end{groupplot}

\end{tikzpicture}

%% file: img/section_06/prediction_03/2d_plots/28.tex
\newcommand{\imgfolder}{img/section_06/prediction_03/2d_plots/}

\begin{tikzpicture}

\definecolor{darkgray176}{RGB}{176,176,176}

\begin{groupplot}[
    group style={
        group size=8 by 1,
        vertical sep=1.9cm,
        horizontal sep=0.5cm
    }
]

\nextgroupplot[
colormap={mymap}{[1pt]
  rgb(0pt)=(0,0,0.5);
  rgb(22pt)=(0,0,1);
  rgb(25pt)=(0,0,1);
  rgb(68pt)=(0,0.86,1);
  rgb(70pt)=(0,0.9,0.967741935483871);
  rgb(75pt)=(0.0806451612903226,1,0.887096774193548);
  rgb(128pt)=(0.935483870967742,1,0.0322580645161291);
  rgb(130pt)=(0.967741935483871,0.962962962962963,0);
  rgb(132pt)=(1,0.925925925925926,0);
  rgb(178pt)=(1,0.0740740740740741,0);
  rgb(182pt)=(0.909090909090909,0,0);
  rgb(200pt)=(0.5,0,0)
},
point meta max=0,
point meta min=-0.795544564723969,
tick align=outside,
x grid style={darkgray176},
xlabel={},
xmajorticks=false,
xmin=-0.5, xmax=319.5,
xtick pos=both,
xtick style={color=black},
y dir=reverse,
y grid style={darkgray176},
ylabel style={rotate=-90.0},
ylabel={Y},
ymin=-0.5, ymax=239.5,
ytick pos=left,
ytick style={color=black}
]
\addplot graphics [includegraphics cmd=\pgfimage,xmin=-0.5, xmax=319.5, ymin=239.5, ymax=-0.5] {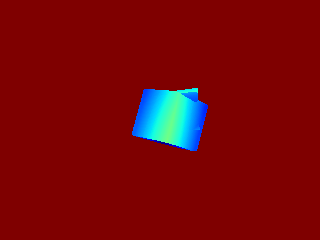};

\nextgroupplot[
colorbar,
colormap={mymap}{[1pt]
  rgb(0pt)=(0,0,0.5);
  rgb(22pt)=(0,0,1);
  rgb(25pt)=(0,0,1);
  rgb(68pt)=(0,0.86,1);
  rgb(70pt)=(0,0.9,0.967741935483871);
  rgb(75pt)=(0.0806451612903226,1,0.887096774193548);
  rgb(128pt)=(0.935483870967742,1,0.0322580645161291);
  rgb(130pt)=(0.967741935483871,0.962962962962963,0);
  rgb(132pt)=(1,0.925925925925926,0);
  rgb(178pt)=(1,0.0740740740740741,0);
  rgb(182pt)=(0.909090909090909,0,0);
  rgb(200pt)=(0.5,0,0)
},
point meta max=0,
point meta min=-0.795544564723969,
tick align=outside,
x grid style={darkgray176},
xlabel={},
xmajorticks=false,
xmin=-0.5, xmax=319.5,
xtick pos=both,
xtick style={color=black},
y dir=reverse,
y grid style={darkgray176},
ylabel style={rotate=-90.0},
ylabel={},
ymajorticks=false,
ymin=-0.5, ymax=239.5,
ytick pos=left,
ytick style={color=black}
]
\addplot graphics [includegraphics cmd=\pgfimage,xmin=-0.5, xmax=319.5, ymin=239.5, ymax=-0.5] {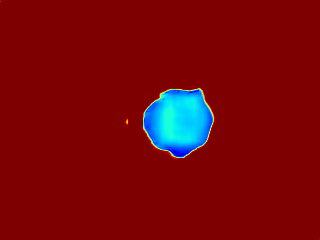};

\nextgroupplot[group/empty plot, width=2.4cm] %

\nextgroupplot[
colormap={mymap}{[1pt]
  rgb(0pt)=(0,0,0.5);
  rgb(22pt)=(0,0,1);
  rgb(25pt)=(0,0,1);
  rgb(68pt)=(0,0.86,1);
  rgb(70pt)=(0,0.9,0.967741935483871);
  rgb(75pt)=(0.0806451612903226,1,0.887096774193548);
  rgb(128pt)=(0.935483870967742,1,0.0322580645161291);
  rgb(130pt)=(0.967741935483871,0.962962962962963,0);
  rgb(132pt)=(1,0.925925925925926,0);
  rgb(178pt)=(1,0.0740740740740741,0);
  rgb(182pt)=(0.909090909090909,0,0);
  rgb(200pt)=(0.5,0,0)
},
point meta max=0.913667142391205,
point meta min=0,
tick align=outside,
x grid style={darkgray176},
xlabel={},
xmajorticks=false,
xmin=-0.5, xmax=319.5,
xtick pos=both,
xtick style={color=black},
y dir=reverse,
y grid style={darkgray176},
ylabel style={rotate=-90.0},
ylabel={},
ymajorticks=false,
ymin=-0.5, ymax=239.5,
ytick pos=left,
ytick style={color=black}
]
\addplot graphics [includegraphics cmd=\pgfimage,xmin=-0.5, xmax=319.5, ymin=239.5, ymax=-0.5] {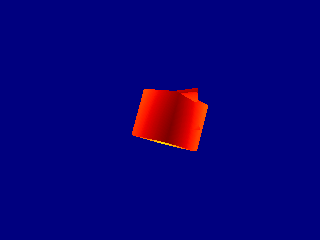};

\nextgroupplot[
colorbar,
colormap={mymap}{[1pt]
  rgb(0pt)=(0,0,0.5);
  rgb(22pt)=(0,0,1);
  rgb(25pt)=(0,0,1);
  rgb(68pt)=(0,0.86,1);
  rgb(70pt)=(0,0.9,0.967741935483871);
  rgb(75pt)=(0.0806451612903226,1,0.887096774193548);
  rgb(128pt)=(0.935483870967742,1,0.0322580645161291);
  rgb(130pt)=(0.967741935483871,0.962962962962963,0);
  rgb(132pt)=(1,0.925925925925926,0);
  rgb(178pt)=(1,0.0740740740740741,0);
  rgb(182pt)=(0.909090909090909,0,0);
  rgb(200pt)=(0.5,0,0)
},
point meta max=0.913667142391205,
point meta min=0,
tick align=outside,
x grid style={darkgray176},
xlabel={},
xmajorticks=false,
xmin=-0.5, xmax=319.5,
xtick pos=both,
xtick style={color=black},
y dir=reverse,
y grid style={darkgray176},
ylabel style={rotate=-90.0},
ylabel={},
ymajorticks=false,
ymin=-0.5, ymax=239.5,
ytick pos=left,
ytick style={color=black}
]
\addplot graphics [includegraphics cmd=\pgfimage,xmin=-0.5, xmax=319.5, ymin=239.5, ymax=-0.5] {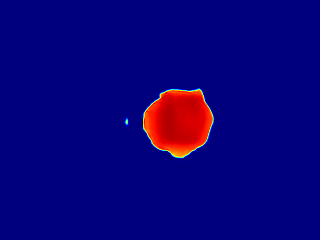};

\nextgroupplot[group/empty plot, width=2.4cm] %

\nextgroupplot[
colormap={mymap}{[1pt]
  rgb(0pt)=(0,0,0.5);
  rgb(22pt)=(0,0,1);
  rgb(25pt)=(0,0,1);
  rgb(68pt)=(0,0.86,1);
  rgb(70pt)=(0,0.9,0.967741935483871);
  rgb(75pt)=(0.0806451612903226,1,0.887096774193548);
  rgb(128pt)=(0.935483870967742,1,0.0322580645161291);
  rgb(130pt)=(0.967741935483871,0.962962962962963,0);
  rgb(132pt)=(1,0.925925925925926,0);
  rgb(178pt)=(1,0.0740740740740741,0);
  rgb(182pt)=(0.909090909090909,0,0);
  rgb(200pt)=(0.5,0,0)
},
point meta max=2.99245081984676,
point meta min=0,
tick align=outside,
x grid style={darkgray176},
xlabel={},
xmajorticks=false,
xmin=-0.5, xmax=319.5,
xtick pos=both,
xtick style={color=black},
y dir=reverse,
y grid style={darkgray176},
ylabel style={rotate=-90.0},
ylabel={},
ymajorticks=false,
ymin=-0.5, ymax=239.5,
ytick pos=left,
ytick style={color=black}
]
\addplot graphics [includegraphics cmd=\pgfimage,xmin=-0.5, xmax=319.5, ymin=239.5, ymax=-0.5] {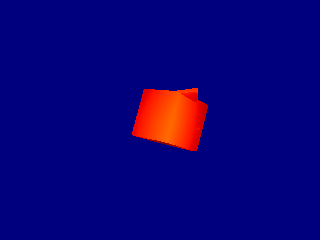};

\nextgroupplot[
colorbar,
colorbar style={
    ylabel={Depth [m]},
    ylabel style={
        yshift=-.5em
    }
},
colormap={mymap}{[1pt]
  rgb(0pt)=(0,0,0.5);
  rgb(22pt)=(0,0,1);
  rgb(25pt)=(0,0,1);
  rgb(68pt)=(0,0.86,1);
  rgb(70pt)=(0,0.9,0.967741935483871);
  rgb(75pt)=(0.0806451612903226,1,0.887096774193548);
  rgb(128pt)=(0.935483870967742,1,0.0322580645161291);
  rgb(130pt)=(0.967741935483871,0.962962962962963,0);
  rgb(132pt)=(1,0.925925925925926,0);
  rgb(178pt)=(1,0.0740740740740741,0);
  rgb(182pt)=(0.909090909090909,0,0);
  rgb(200pt)=(0.5,0,0)
},
point meta max=2.99245081984676,
point meta min=0,
tick align=outside,
x grid style={darkgray176},
xlabel={},
xmajorticks=false,
xmin=-0.5, xmax=319.5,
xtick pos=both,
xtick style={color=black},
y dir=reverse,
y grid style={darkgray176},
ylabel style={rotate=-90.0},
ylabel={},
ymajorticks=false,
ymin=-0.5, ymax=239.5,
ytick pos=left,
ytick style={color=black}
]
\addplot graphics [includegraphics cmd=\pgfimage,xmin=-0.5, xmax=319.5, ymin=239.5, ymax=-0.5] {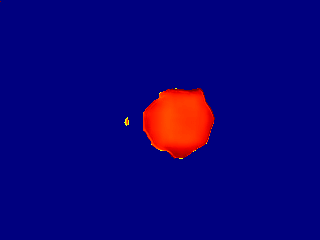};
\draw (axis cs:15,15) node[
  scale=1.5,
  fill=white,
  draw=black,
  line width=0.4pt,
  inner sep=3pt,
  fill opacity=0.9,
  rounded corners=0.5mm,
  anchor=north west,
  text=black,
  rotate=0.0
]{MAE: $5.1$cm};
\draw (axis cs:15,45) node[
  scale=1.5,
  fill=white,
  draw=black,
  line width=0.4pt,
  inner sep=3pt,
  fill opacity=0.9,
  rounded corners=0.5mm,
  anchor=north west,
  text=black,
  rotate=0.0
]{mIoU: $0.837$};

\end{groupplot}

\end{tikzpicture}

%% file: img/section_06/prediction_04/2d_plots/74.tex
\newcommand{\imgfolder}{img/section_06/prediction_04/2d_plots/}

\begin{tikzpicture}

\definecolor{darkgray176}{RGB}{176,176,176}

\begin{groupplot}[
    group style={
        group size=8 by 1,
        vertical sep=1.9cm,
        horizontal sep=0.5cm
    }
]

\nextgroupplot[
colormap={mymap}{[1pt]
  rgb(0pt)=(0,0,0.5);
  rgb(22pt)=(0,0,1);
  rgb(25pt)=(0,0,1);
  rgb(68pt)=(0,0.86,1);
  rgb(70pt)=(0,0.9,0.967741935483871);
  rgb(75pt)=(0.0806451612903226,1,0.887096774193548);
  rgb(128pt)=(0.935483870967742,1,0.0322580645161291);
  rgb(130pt)=(0.967741935483871,0.962962962962963,0);
  rgb(132pt)=(1,0.925925925925926,0);
  rgb(178pt)=(1,0.0740740740740741,0);
  rgb(182pt)=(0.909090909090909,0,0);
  rgb(200pt)=(0.5,0,0)
},
point meta max=0,
point meta min=-0.752599239349365,
tick align=outside,
x grid style={darkgray176},
xlabel={},
xmajorticks=false,
xmin=-0.5, xmax=319.5,
xtick pos=bottom,
xtick style={color=black},
y dir=reverse,
y grid style={darkgray176},
ylabel style={rotate=-90.0},
ylabel={Y},
ymajorticks=true,
ymin=-0.5, ymax=239.5,
ytick pos=left,
ytick style={color=black}
]
\addplot graphics [includegraphics cmd=\pgfimage,xmin=-0.5, xmax=319.5, ymin=239.5, ymax=-0.5] {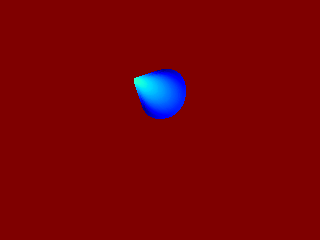};

\nextgroupplot[
colorbar,
colorbar style={
    ylabel style={
        yshift=-9em
    }
},
colormap={mymap}{[1pt]
  rgb(0pt)=(0,0,0.5);
  rgb(22pt)=(0,0,1);
  rgb(25pt)=(0,0,1);
  rgb(68pt)=(0,0.86,1);
  rgb(70pt)=(0,0.9,0.967741935483871);
  rgb(75pt)=(0.0806451612903226,1,0.887096774193548);
  rgb(128pt)=(0.935483870967742,1,0.0322580645161291);
  rgb(130pt)=(0.967741935483871,0.962962962962963,0);
  rgb(132pt)=(1,0.925925925925926,0);
  rgb(178pt)=(1,0.0740740740740741,0);
  rgb(182pt)=(0.909090909090909,0,0);
  rgb(200pt)=(0.5,0,0)
},
point meta max=0,
point meta min=-0.752599239349365,
tick align=outside,
x grid style={darkgray176},
xlabel={},
xmajorticks=false,
xmin=-0.5, xmax=319.5,
xtick pos=bottom,
xtick style={color=black},
y dir=reverse,
y grid style={darkgray176},
ylabel style={rotate=-90.0},
ylabel={},
ymajorticks=false,
ymin=-0.5, ymax=239.5,
ytick pos=left,
ytick style={color=black}
]
\addplot graphics [includegraphics cmd=\pgfimage,xmin=-0.5, xmax=319.5, ymin=239.5, ymax=-0.5] {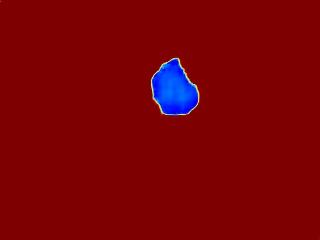};

\nextgroupplot[group/empty plot, width=2.4cm] %

\nextgroupplot[
colormap={mymap}{[1pt]
  rgb(0pt)=(0,0,0.5);
  rgb(22pt)=(0,0,1);
  rgb(25pt)=(0,0,1);
  rgb(68pt)=(0,0.86,1);
  rgb(70pt)=(0,0.9,0.967741935483871);
  rgb(75pt)=(0.0806451612903226,1,0.887096774193548);
  rgb(128pt)=(0.935483870967742,1,0.0322580645161291);
  rgb(130pt)=(0.967741935483871,0.962962962962963,0);
  rgb(132pt)=(1,0.925925925925926,0);
  rgb(178pt)=(1,0.0740740740740741,0);
  rgb(182pt)=(0.909090909090909,0,0);
  rgb(200pt)=(0.5,0,0)
},
point meta max=0.895085513591766,
point meta min=0,
tick align=outside,
x grid style={darkgray176},
xlabel={},
xmajorticks=false,
xmin=-0.5, xmax=319.5,
xtick pos=bottom,
xtick style={color=black},
y dir=reverse,
y grid style={darkgray176},
ylabel style={rotate=-90.0},
ylabel={},
ymajorticks=false,
ymin=-0.5, ymax=239.5,
ytick pos=left,
ytick style={color=black}
]
\addplot graphics [includegraphics cmd=\pgfimage,xmin=-0.5, xmax=319.5, ymin=239.5, ymax=-0.5] {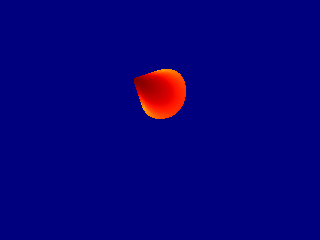};

\nextgroupplot[
colorbar,
colorbar style={
    ylabel style={
        yshift=-9em
    }
},
colormap={mymap}{[1pt]
  rgb(0pt)=(0,0,0.5);
  rgb(22pt)=(0,0,1);
  rgb(25pt)=(0,0,1);
  rgb(68pt)=(0,0.86,1);
  rgb(70pt)=(0,0.9,0.967741935483871);
  rgb(75pt)=(0.0806451612903226,1,0.887096774193548);
  rgb(128pt)=(0.935483870967742,1,0.0322580645161291);
  rgb(130pt)=(0.967741935483871,0.962962962962963,0);
  rgb(132pt)=(1,0.925925925925926,0);
  rgb(178pt)=(1,0.0740740740740741,0);
  rgb(182pt)=(0.909090909090909,0,0);
  rgb(200pt)=(0.5,0,0)
},
point meta max=0.895085513591766,
point meta min=0,
tick align=outside,
x grid style={darkgray176},
xlabel={},
xmajorticks=false,
xmin=-0.5, xmax=319.5,
xtick pos=bottom,
xtick style={color=black},
y dir=reverse,
y grid style={darkgray176},
ylabel style={rotate=-90.0},
ylabel={},
ymajorticks=false,
ymin=-0.5, ymax=239.5,
ytick pos=left,
ytick style={color=black}
]
\addplot graphics [includegraphics cmd=\pgfimage,xmin=-0.5, xmax=319.5, ymin=239.5, ymax=-0.5] {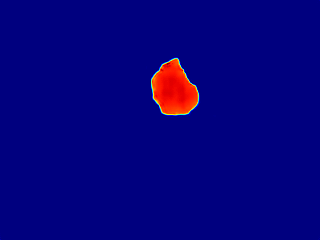};

\nextgroupplot[group/empty plot, width=2.4cm] %

\nextgroupplot[
colormap={mymap}{[1pt]
  rgb(0pt)=(0,0,0.5);
  rgb(22pt)=(0,0,1);
  rgb(25pt)=(0,0,1);
  rgb(68pt)=(0,0.86,1);
  rgb(70pt)=(0,0.9,0.967741935483871);
  rgb(75pt)=(0.0806451612903226,1,0.887096774193548);
  rgb(128pt)=(0.935483870967742,1,0.0322580645161291);
  rgb(130pt)=(0.967741935483871,0.962962962962963,0);
  rgb(132pt)=(1,0.925925925925926,0);
  rgb(178pt)=(1,0.0740740740740741,0);
  rgb(182pt)=(0.909090909090909,0,0);
  rgb(200pt)=(0.5,0,0)
},
point meta max=2.97292562008537,
point meta min=0,
tick align=outside,
x grid style={darkgray176},
xlabel={},
xmajorticks=false,
xmin=-0.5, xmax=319.5,
xtick pos=bottom,
xtick style={color=black},
y dir=reverse,
y grid style={darkgray176},
ylabel style={rotate=-90.0},
ylabel={},
ymajorticks=false,
ymin=-0.5, ymax=239.5,
ytick pos=left,
ytick style={color=black}
]
\addplot graphics [includegraphics cmd=\pgfimage,xmin=-0.5, xmax=319.5, ymin=239.5, ymax=-0.5] {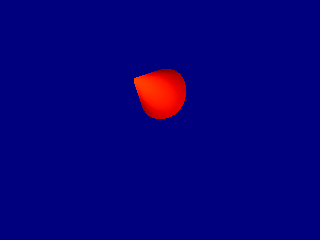};

\nextgroupplot[
colorbar,
colorbar style={
    ylabel={Depth [m]},
    ylabel style={
        yshift=-.5em
    }
},
colormap={mymap}{[1pt]
  rgb(0pt)=(0,0,0.5);
  rgb(22pt)=(0,0,1);
  rgb(25pt)=(0,0,1);
  rgb(68pt)=(0,0.86,1);
  rgb(70pt)=(0,0.9,0.967741935483871);
  rgb(75pt)=(0.0806451612903226,1,0.887096774193548);
  rgb(128pt)=(0.935483870967742,1,0.0322580645161291);
  rgb(130pt)=(0.967741935483871,0.962962962962963,0);
  rgb(132pt)=(1,0.925925925925926,0);
  rgb(178pt)=(1,0.0740740740740741,0);
  rgb(182pt)=(0.909090909090909,0,0);
  rgb(200pt)=(0.5,0,0)
},
point meta max=2.97292562008537,
point meta min=0,
tick align=outside,
x grid style={darkgray176},
xlabel={},
xmajorticks=false,
xmin=-0.5, xmax=319.5,
xtick pos=bottom,
xtick style={color=black},
y dir=reverse,
y grid style={darkgray176},
ylabel style={rotate=-90.0},
ylabel={},
ymajorticks=false,
ymin=-0.5, ymax=239.5,
ytick pos=left,
ytick style={color=black}
]
\addplot graphics [includegraphics cmd=\pgfimage,xmin=-0.5, xmax=319.5, ymin=239.5, ymax=-0.5] {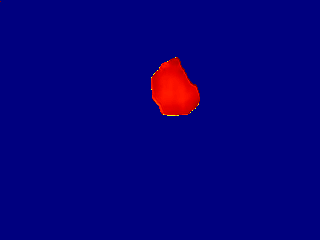};
\draw (axis cs:15,15) node[
  scale=1.5,
  fill=white,
  draw=black,
  line width=0.4pt,
  inner sep=3pt,
  fill opacity=0.9,
  rounded corners=0.5mm,
  anchor=north west,
  text=black,
  rotate=0.0
]{MAE: $5.1$cm};
\draw (axis cs:15,45) node[
  scale=1.5,
  fill=white,
  draw=black,
  line width=0.4pt,
  inner sep=3pt,
  fill opacity=0.9,
  rounded corners=0.5mm,
  anchor=north west,
  text=black,
  rotate=0.0
]{mIoU: $0.729$};

\end{groupplot}

\end{tikzpicture}

%% file: img/section_06/prediction_05/2d_plots/31.tex
\newcommand{\imgfolder}{img/section_06/prediction_05/2d_plots/}

\begin{tikzpicture}

\definecolor{darkgray176}{RGB}{176,176,176}

\begin{groupplot}[
    group style={
        group size=8 by 1,
        vertical sep=1.9cm,
        horizontal sep=0.5cm
    }
]

\nextgroupplot[
colormap={mymap}{[1pt]
  rgb(0pt)=(0,0,0.5);
  rgb(22pt)=(0,0,1);
  rgb(25pt)=(0,0,1);
  rgb(68pt)=(0,0.86,1);
  rgb(70pt)=(0,0.9,0.967741935483871);
  rgb(75pt)=(0.0806451612903226,1,0.887096774193548);
  rgb(128pt)=(0.935483870967742,1,0.0322580645161291);
  rgb(130pt)=(0.967741935483871,0.962962962962963,0);
  rgb(132pt)=(1,0.925925925925926,0);
  rgb(178pt)=(1,0.0740740740740741,0);
  rgb(182pt)=(0.909090909090909,0,0);
  rgb(200pt)=(0.5,0,0)
},
point meta max=0,
point meta min=-0.723187863826752,
tick align=outside,
x grid style={darkgray176},
xlabel={X},
xmin=-0.5, xmax=319.5,
xtick pos=bottom,
xtick style={color=black},
y dir=reverse,
y grid style={darkgray176},
ylabel style={rotate=-90.0},
ylabel={Y},
ymin=-0.5, ymax=239.5,
ytick pos=left,
ytick style={color=black}
]
\addplot graphics [includegraphics cmd=\pgfimage,xmin=-0.5, xmax=319.5, ymin=239.5, ymax=-0.5] {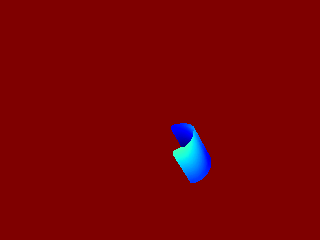};

\nextgroupplot[
colorbar,
colormap={mymap}{[1pt]
  rgb(0pt)=(0,0,0.5);
  rgb(22pt)=(0,0,1);
  rgb(25pt)=(0,0,1);
  rgb(68pt)=(0,0.86,1);
  rgb(70pt)=(0,0.9,0.967741935483871);
  rgb(75pt)=(0.0806451612903226,1,0.887096774193548);
  rgb(128pt)=(0.935483870967742,1,0.0322580645161291);
  rgb(130pt)=(0.967741935483871,0.962962962962963,0);
  rgb(132pt)=(1,0.925925925925926,0);
  rgb(178pt)=(1,0.0740740740740741,0);
  rgb(182pt)=(0.909090909090909,0,0);
  rgb(200pt)=(0.5,0,0)
},
point meta max=0,
point meta min=-0.723187863826752,
tick align=outside,
x grid style={darkgray176},
xlabel={X},
xmin=-0.5, xmax=319.5,
xtick pos=both,
xtick style={color=black},
y dir=reverse,
y grid style={darkgray176},
ylabel style={rotate=-90.0},
ylabel={},
ymajorticks=false,
ymin=-0.5, ymax=239.5,
ytick pos=left,
ytick style={color=black}
]
\addplot graphics [includegraphics cmd=\pgfimage,xmin=-0.5, xmax=319.5, ymin=239.5, ymax=-0.5] {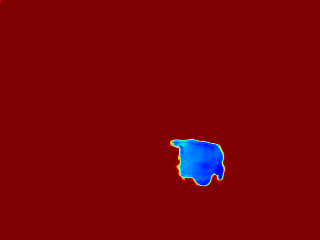};

\nextgroupplot[group/empty plot, width=2.4cm] %

\nextgroupplot[
colormap={mymap}{[1pt]
  rgb(0pt)=(0,0,0.5);
  rgb(22pt)=(0,0,1);
  rgb(25pt)=(0,0,1);
  rgb(68pt)=(0,0.86,1);
  rgb(70pt)=(0,0.9,0.967741935483871);
  rgb(75pt)=(0.0806451612903226,1,0.887096774193548);
  rgb(128pt)=(0.935483870967742,1,0.0322580645161291);
  rgb(130pt)=(0.967741935483871,0.962962962962963,0);
  rgb(132pt)=(1,0.925925925925926,0);
  rgb(178pt)=(1,0.0740740740740741,0);
  rgb(182pt)=(0.909090909090909,0,0);
  rgb(200pt)=(0.5,0,0)
},
point meta max=0.910227537155151,
point meta min=0,
tick align=outside,
x grid style={darkgray176},
xlabel={X},
xmin=-0.5, xmax=319.5,
xtick pos=bottom,
xtick style={color=black},
y dir=reverse,
y grid style={darkgray176},
ylabel style={rotate=-90.0},
ylabel={},
ymajorticks=false,
ymin=-0.5, ymax=239.5,
ytick pos=left,
ytick style={color=black}
]
\addplot graphics [includegraphics cmd=\pgfimage,xmin=-0.5, xmax=319.5, ymin=239.5, ymax=-0.5] {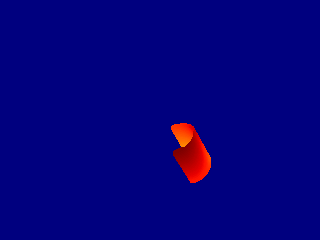};

\nextgroupplot[
colorbar,
colormap={mymap}{[1pt]
  rgb(0pt)=(0,0,0.5);
  rgb(22pt)=(0,0,1);
  rgb(25pt)=(0,0,1);
  rgb(68pt)=(0,0.86,1);
  rgb(70pt)=(0,0.9,0.967741935483871);
  rgb(75pt)=(0.0806451612903226,1,0.887096774193548);
  rgb(128pt)=(0.935483870967742,1,0.0322580645161291);
  rgb(130pt)=(0.967741935483871,0.962962962962963,0);
  rgb(132pt)=(1,0.925925925925926,0);
  rgb(178pt)=(1,0.0740740740740741,0);
  rgb(182pt)=(0.909090909090909,0,0);
  rgb(200pt)=(0.5,0,0)
},
point meta max=0.910227537155151,
point meta min=0,
tick align=outside,
x grid style={darkgray176},
xlabel={X},
xmin=-0.5, xmax=319.5,
xtick pos=bottom,
xtick style={color=black},
y dir=reverse,
y grid style={darkgray176},
ylabel style={rotate=-90.0},
ylabel={},
ymajorticks=false,
ymin=-0.5, ymax=239.5,
ytick pos=left,
ytick style={color=black}
]
\addplot graphics [includegraphics cmd=\pgfimage,xmin=-0.5, xmax=319.5, ymin=239.5, ymax=-0.5] {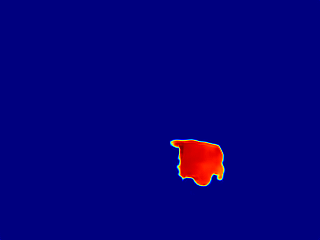};

\nextgroupplot[group/empty plot, width=2.4cm] %

\nextgroupplot[
colormap={mymap}{[1pt]
  rgb(0pt)=(0,0,0.5);
  rgb(22pt)=(0,0,1);
  rgb(25pt)=(0,0,1);
  rgb(68pt)=(0,0.86,1);
  rgb(70pt)=(0,0.9,0.967741935483871);
  rgb(75pt)=(0.0806451612903226,1,0.887096774193548);
  rgb(128pt)=(0.935483870967742,1,0.0322580645161291);
  rgb(130pt)=(0.967741935483871,0.962962962962963,0);
  rgb(132pt)=(1,0.925925925925926,0);
  rgb(178pt)=(1,0.0740740740740741,0);
  rgb(182pt)=(0.909090909090909,0,0);
  rgb(200pt)=(0.5,0,0)
},
point meta max=2.8895683210952,
point meta min=0,
tick align=outside,
x grid style={darkgray176},
xlabel={X},
xmin=-0.5, xmax=319.5,
xtick pos=bottom,
xtick style={color=black},
y dir=reverse,
y grid style={darkgray176},
ylabel style={rotate=-90.0},
ylabel={},
ymajorticks=false,
ymin=-0.5, ymax=239.5,
ytick pos=left,
ytick style={color=black}
]
\addplot graphics [includegraphics cmd=\pgfimage,xmin=-0.5, xmax=319.5, ymin=239.5, ymax=-0.5] {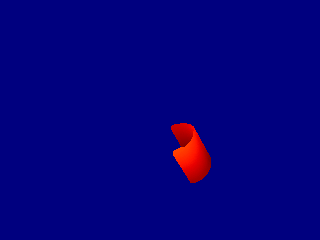};

\nextgroupplot[
colorbar,
colorbar style={
    ylabel={Depth [m]},
    ylabel style={
        yshift=-.5em
    }
},
colormap={mymap}{[1pt]
  rgb(0pt)=(0,0,0.5);
  rgb(22pt)=(0,0,1);
  rgb(25pt)=(0,0,1);
  rgb(68pt)=(0,0.86,1);
  rgb(70pt)=(0,0.9,0.967741935483871);
  rgb(75pt)=(0.0806451612903226,1,0.887096774193548);
  rgb(128pt)=(0.935483870967742,1,0.0322580645161291);
  rgb(130pt)=(0.967741935483871,0.962962962962963,0);
  rgb(132pt)=(1,0.925925925925926,0);
  rgb(178pt)=(1,0.0740740740740741,0);
  rgb(182pt)=(0.909090909090909,0,0);
  rgb(200pt)=(0.5,0,0)
},
point meta max=2.8895683210952,
point meta min=0,
tick align=outside,
x grid style={darkgray176},
xlabel={X},
xmin=-0.5, xmax=319.5,
xtick pos=bottom,
xtick style={color=black},
y dir=reverse,
y grid style={darkgray176},
ylabel style={rotate=-90.0},
ylabel={},
ymajorticks=false,
ymin=-0.5, ymax=239.5,
ytick pos=left,
ytick style={color=black}
]
\addplot graphics [includegraphics cmd=\pgfimage,xmin=-0.5, xmax=319.5, ymin=239.5, ymax=-0.5] {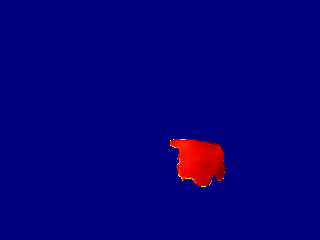};
\draw (axis cs:15,15) node[
  scale=1.5,
  fill=white,
  draw=black,
  line width=0.4pt,
  inner sep=3pt,
  fill opacity=0.9,
  rounded corners=0.5,
  anchor=north west,
  text=black,
  rotate=0.0
]{MAE: $4.5$cm};
\draw (axis cs:15,45) node[
  scale=1.5,
  fill=white,
  draw=black,
  line width=0.4pt,
  inner sep=3pt,
  fill opacity=0.9,
  rounded corners=0.5,
  anchor=north west,
  text=black,
  rotate=0.0
]{mIoU: $0.725$};

\end{groupplot}

\end{tikzpicture}

%% file: img/section_06/pc.tex
\begin{figure}[!htb]
    \centering
    \subfloat{
        \includegraphics[width=.14\linewidth]{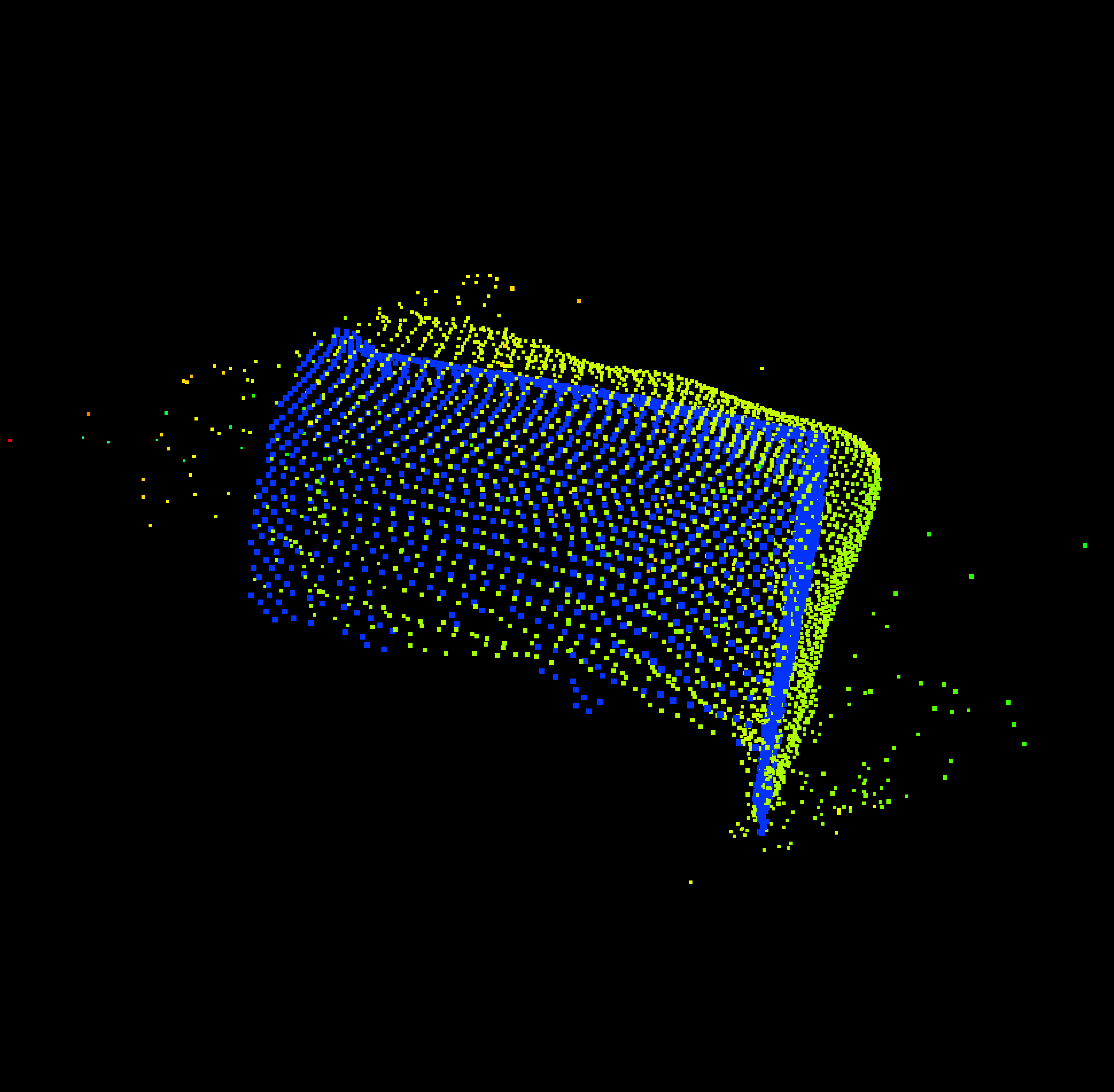}
    }
    \,
    \subfloat{
        \includegraphics[width=.14\linewidth]{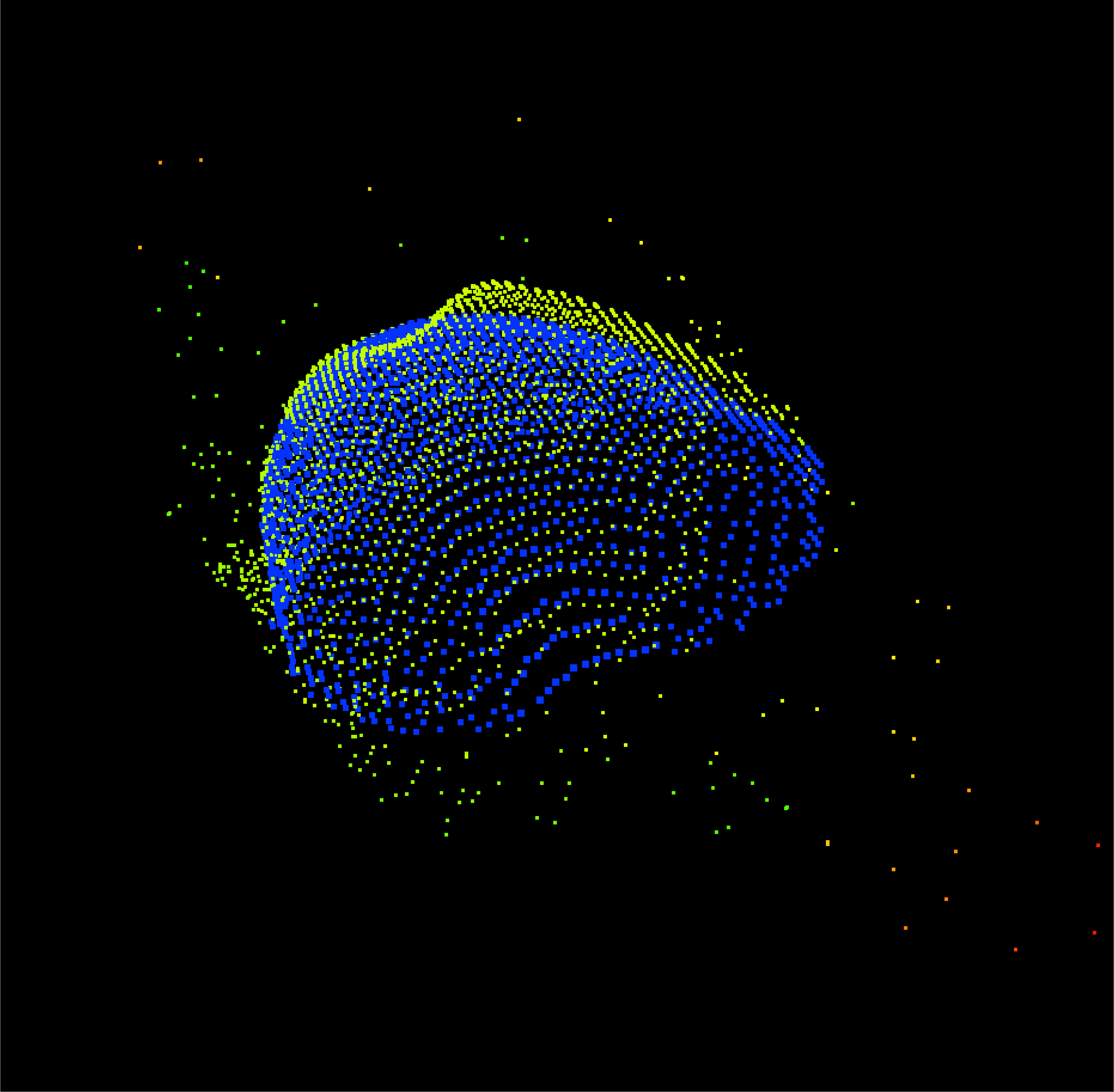}
    }
    \,
    \subfloat{
        \includegraphics[width=.14\linewidth]{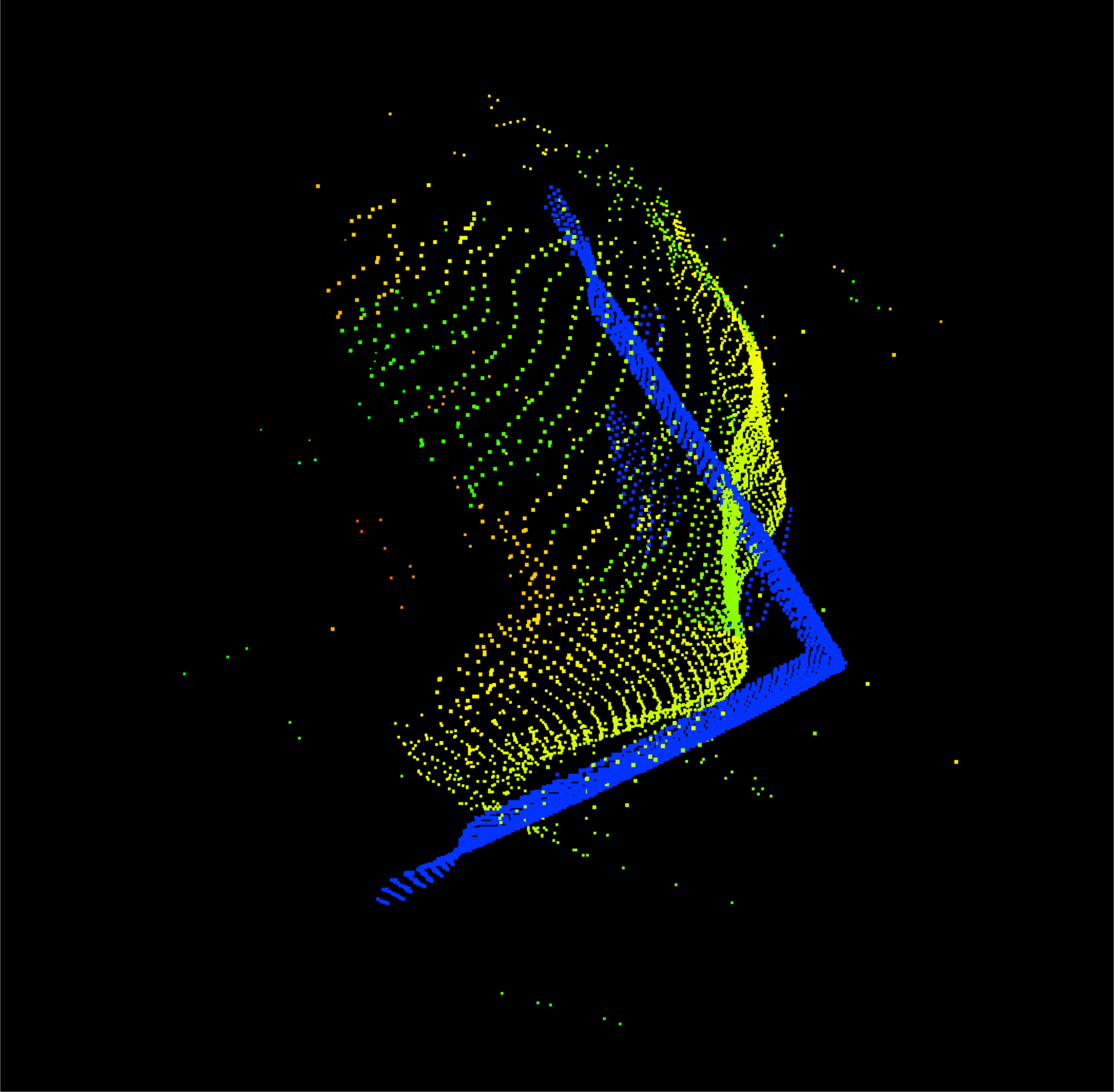}
    }
    \,
    \subfloat{
        \includegraphics[width=.14\linewidth]{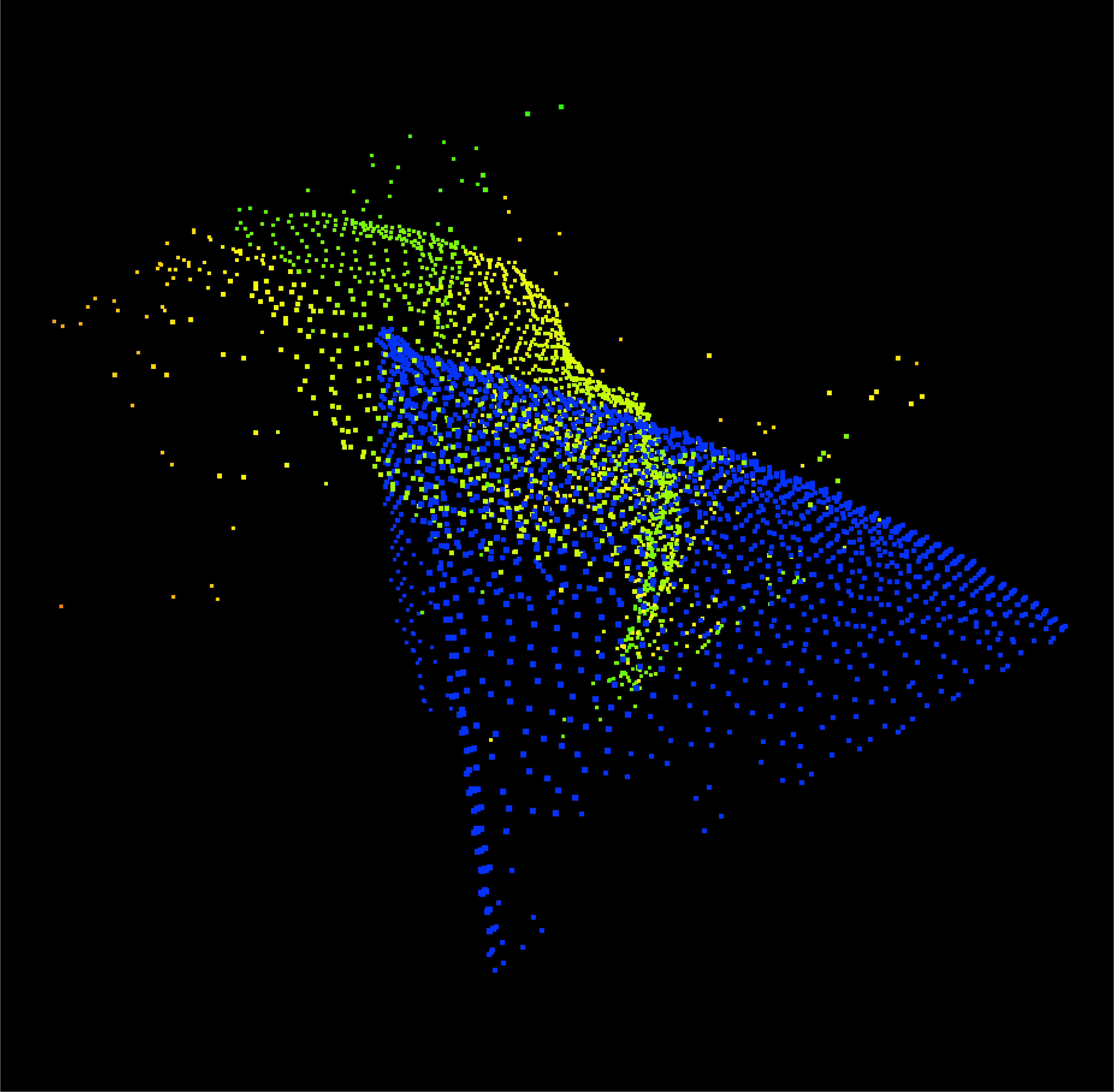}
    }
    \,
    \subfloat{
        \includegraphics[width=.14\linewidth]{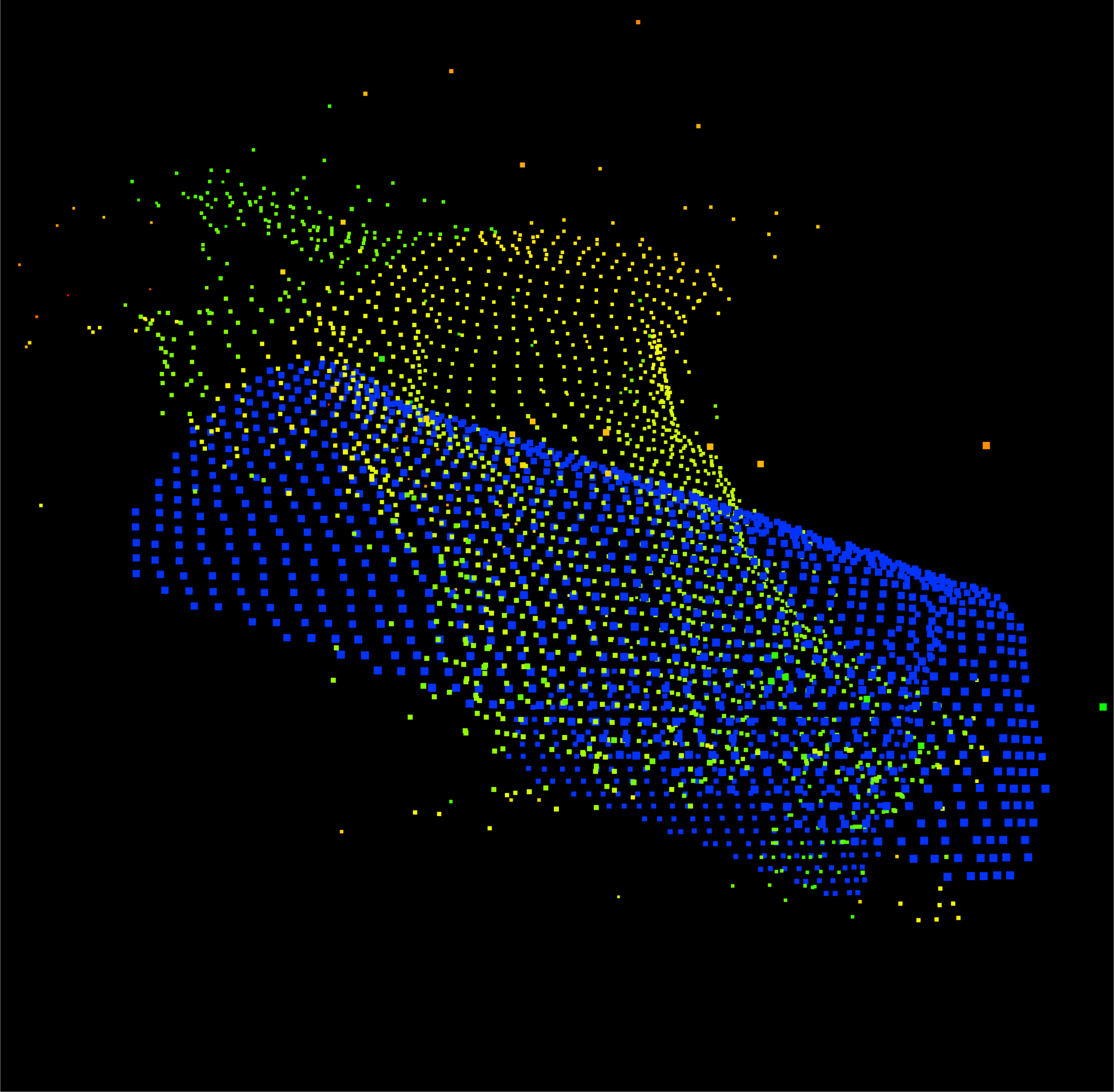}
    }
    \caption{Point cloud examples (\lbrack{\color{blue} $\bullet$}\rbrack$=$\acs{gt}, \lbrack{\color{green} $\bullet$} $\rightarrow$ {\color{red} $\bullet$}\rbrack$=$\lbrack closer $\rightarrow$ further\rbrack$=$prediction)}
    \label{fig:pc}
\end{figure}

%% file: sec/07_limitations.tex
\section{Limitations} \label{sec:limitations}
    The limitations of this approach are primarily associated with the transition from synthetic to real datasets, which requires a very accurate model of the sensor noise to ensure reliable results. Additionally, the overall precision of the proposed model is still limited, as it struggles to accurately recover complex geometries due to the significant signal degradation caused by multiple reflections.

%% file: sec/08_conclusions.tex
\section{Conclusions} \label{sec:conclusions}
    We tackled the task of \acl{nlos} imaging using an \acl{itof} sensor as the source of measurements: the proposed approach that combines the \textit{Mirror trick} with an ad-hoc \acl{dl} solution proved to be a feasible solution for the task. The use of this specific type of \ac{tof} cameras, instead of the more common \ac{dtof}, allows us to use off-the-shelf cameras, greatly reducing the cost and complexity of the system. Furthermore, by performing a \textit{full field} illumination, the acquisition becomes extremely fast (multiple frames each second), allowing real-time applications. This, together with the good reconstruction results, makes the proposed approach an interesting alternative for \acl{nlos} imaging. A key future step is to assess the approach with real-world data. The \textit{Mirror trick} could be used to generate real-world \acl{gt} by replacing the front wall with a mirror and reacquiring the scene.

%% file: sec/09_acknowledgment.tex
\section*{Acknowledgment} \label{sec:acknowledgment}
    Thanks to \href{https://scholar.google.com/citations?user=Ysz_fQoAAAAJ}{Gianluca Agresti} for introducing some of the presented ideas and for supporting this work.
    
    This work was supported in part by the European Union through the Italian National Recovery and Resilience Plan (NRRP) of NextGenerationEU, a partnership on “Telecommunications of the Future” (Program “RE\-START”) under Grant PE0000001.